\documentclass{article}

\usepackage{omniweaving_arxiv,times}

\usepackage[utf8]{inputenc} 
\usepackage[T1]{fontenc}    
\makeatletter
\@ifpackageloaded{hyperref}{}{\usepackage{hyperref}} 
\@ifpackageloaded{url}{}{\usepackage{url}}           
\makeatother 
\usepackage{booktabs}       
\usepackage{array}
\usepackage{caption}
\usepackage{amsmath}
\usepackage{amssymb}
\usepackage{amsthm}
\usepackage{amsfonts}       
\usepackage{nicefrac}       
\usepackage{microtype}      
\usepackage{xcolor}         
\usepackage{float}         
\usepackage{wrapfig}
\usepackage{pgfplots}
\usepackage{graphicx}
\usepackage{subcaption}
\usepackage{multirow}
\usepackage{booktabs} 
\pgfplotsset{compat=1.18}
\usepgfplotslibrary{groupplots}

\newcommand{\condvspace}[1]{}
\renewcommand{\cite}{\citep}

\makeatletter
\@ifundefined{c@theorem}{}{}
\@ifundefined{c@lemma}{}{}
\@ifundefined{c@proposition}{}{}
\@ifundefined{c@corollary}{}{}
\theoremstyle{definition}
\@ifundefined{c@assumption}{}{}
\@ifundefined{c@construction}{}{}
\makeatother

\title{TAGRPO: Boosting GRPO on Image-to-Video Generation with Direct Trajectory Alignment} 

\author{%
  Jin Wang$^{1,2,*}$, Jianxiang Lu$^{2,*}$, Guangzheng Xu$^{2}$, Comi Chen$^{2}$, \\
  Haoyu Yang$^{2}$, Linqing Wang$^{2}$, Peng Chen$^{2}$,
  Mingtao Chen$^{2}$, Zhichao Hu$^{2}$, Longhuang Wu$^{2,\dagger}$, Shuai Shao$^{2}$, Qinglin Lu$^{2}$, Ping Luo$^{1,\S}$\\[2pt]
  $^{1}$The University of Hong Kong \quad $^{2}$Tencent Hunyuan\\
  $^{*}$ Equal Contribution, $^{\dagger}$ Project Lead, $^{\S}$ Corresponding Author 
}

\begin{document}

\maketitle

\newcommand{\ignore}[1]{}

\begin{abstract}
 Recent studies have demonstrated the efficacy of integrating Group Relative Policy Optimization (GRPO) into flow matching models, particularly for text-to-image and text-to-video generation. However, we find that directly applying these techniques to image-to-video (I2V) models often fails to yield consistent reward improvements. To address this limitation, we present TAGRPO, a robust post-training framework for I2V models inspired by contrastive learning. Our approach is grounded in the observation that rollout videos generated from identical initial noise provide superior guidance for optimization. Leveraging this insight, we propose a novel GRPO loss applied to intermediate latents, encouraging direct alignment with high-reward trajectories while maximizing distance from low-reward counterparts. Furthermore, we introduce a memory bank for rollout videos to enhance diversity and reduce computational overhead. Despite its simplicity, TAGRPO achieves significant improvements over DanceGRPO in I2V generation. The deliverables will be updated \href{https://tagrpo.github.io/}{here}.
\end{abstract}

\section{Introduction}
\label{sec:intro}

With the development of diffusion models \cite{ho2020denoising,songscore,lipmanflow,liuflow,peebles2023scalable,dhariwal2021diffusion}, recent years have witnessed the success of AIGC technology in text-to-image generation \cite{rombach2022high,esser2024scaling,flux2024,labs2025flux1kontextflowmatching,chenpixart} and text-to-video generation \cite{ho2022video,blattmann2023stable,yang2024cogvideox,kong2024hunyuanvideo,wan2025wan,zheng2024open,lin2024open}. To further enhance alignment between generated content and human preferences, recent studies \cite{liu2025flow,xue2025dancegrpo} have applied reinforcement learning techniques, such as GRPO \cite{shao2024deepseekmath}, to visual generative models, achieving significant progress.

Most existing work \cite{he2025tempflow,li2025branchgrpo,li2025mixgrpo,fu2025dynamic} has primarily focused on text-conditioned generation paradigms. In contrast, image-to-video generation \cite{wan2025wan,kong2024hunyuanvideo,chen2025skyreels} remains underexplored, despite its broad applicability in domains such as animation \cite{hu2024animate}, content creation \cite{yang2024hi3d}, and visual effects \cite{mao2025omni}. Notably, we observe that directly applying existing visual GRPO methods \cite{liu2025flow,xue2025dancegrpo} to state-of-the-art image-to-video models—such as Wan 2.2 \cite{wan2025wan} and HunyuanVideo-1.5 \cite{wu2025hunyuanvideo}—fails to yield consistent reward improvements. This observation raises a critical question: \textit{Can we devise an effective GRPO framework tailored for image-to-video generation?}

In this paper, we present TAGRPO, an effective GRPO framework for post-training image-to-video models based on the concept of \textbf{T}rajectory \textbf{A}lignment. We observe that existing methods \cite{liu2025flow,xue2025dancegrpo} typically rely on reward signals to modulate the probability of each sample \textit{individually}, thereby overlooking valuable relational guidance among generated samples within a group. This oversight is critical: since videos generated from the same conditioning image share significant structural content, the \textit{relative relationships} among their trajectories offer rich optimization cues. Consequently, rather than merely suppressing the generation probability of a negative sample, it is more intuitive and effective to further align its trajectory with those of positive samples within the same group.

\begin{figure*}[t]
    \centering
    \begin{subfigure}[t]{0.49\textwidth}
        \centering
        \includegraphics[width=\textwidth]{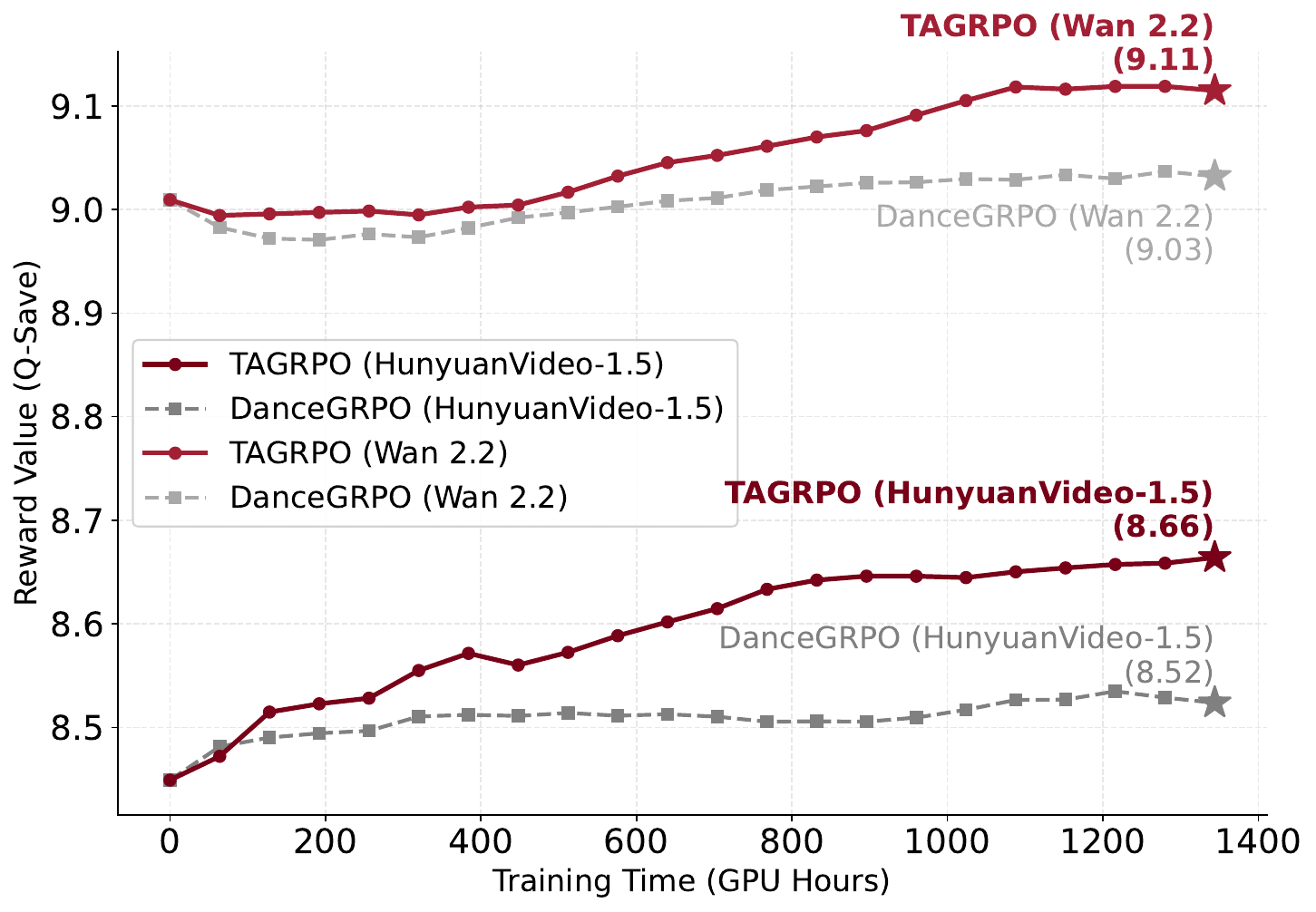}
        \label{fig:performance-qsave}
    \end{subfigure}
    \hfill
    \begin{subfigure}[t]{0.49\textwidth}
        \centering
        \includegraphics[width=\textwidth]{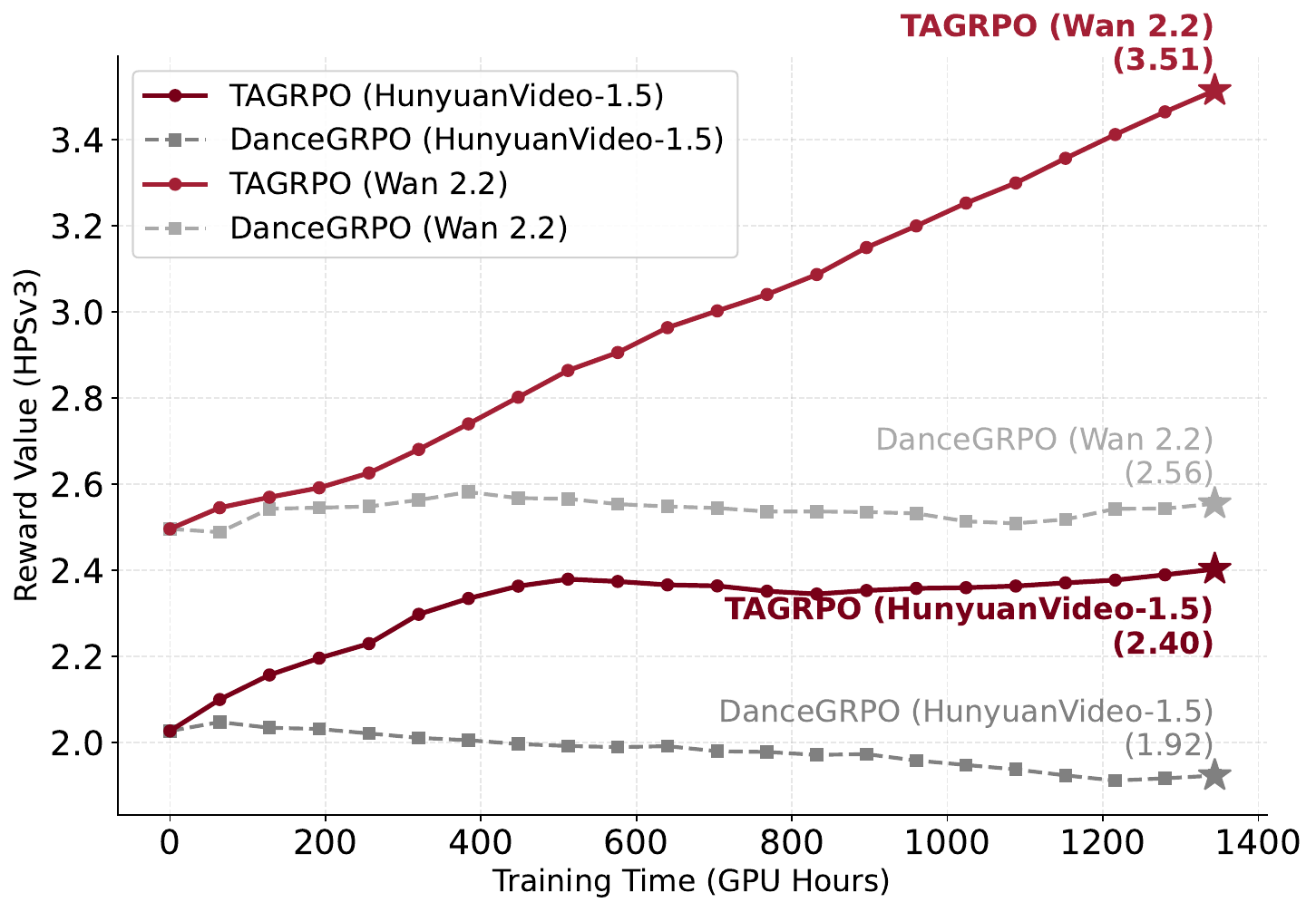}
        \label{fig:performance-hpsv3}
    \end{subfigure}
    \caption{\textbf{Performance of the proposed TAGRPO.} 
    We mainly compared our method with DanceGRPO~\cite{xue2025dancegrpo}, as existing open‑sourced implementations of visual GRPO methods \cite{liu2025flow,he2025tempflow,zheng2025diffusionnft} typically support text‑conditioned tasks, with DanceGRPO being the only exception. 
    The results demonstrate that TAGRPO achieved faster convergence and consistently higher reward gains on both Wan 2.2 \cite{wan2025wan} and HunyuanVideo-1.5 \cite{wu2025hunyuanvideo}. We used Q‑Save~\cite{wu2025q} and HPSv3~\cite{ma2025hpsv3} as reward models, and all reported reward values were averaged over the evaluation set.}
    \label{fig:teaser}
\end{figure*}

To leverage this insight, we propose to directly align the inference trajectory by applying a new trajectory-wise GRPO loss to intermediate latents based on reward rankings. Concretely, we encourage latents to align more closely with those from higher-reward videos while maintaining greater distance from lower-reward counterparts. Experiments demonstrate that this simple yet effective approach yields significant improvements, validating the importance of exploiting inter-sample relationships for image-to-video generation.
Besides, inspired by the core concepts in contrastive learning \cite{he2020momentum}, we propose to maintain a memory bank for keeping previous generated samples' latents and reward signals in our proposed TAGRPO.
This could release the burden for preparing a large batch of rollout videos for every step, allowing the model to effectively exploit previous generated samples.
As shown in Figure \ref{fig:teaser}, we applied our method to advanced image-to-video models \cite{wan2025wan,wu2025hunyuanvideo}, achieving significant improvements over DanceGRPO \cite{xue2025dancegrpo}.

The contributions of our paper are summarized as follows:
1) We propose TAGRPO, a novel trajectory alignment framework that leverages relative relationships among generated samples. This approach provides more informative optimization signals for image-to-video generation.
2) We introduce a memory bank mechanism that enables efficient exploitation of historical samples, significantly reducing the computational requirements for rollout generation while maintaining optimization effectiveness.
3) Through extensive experiments on advanced image-to-video models \cite{wan2025wan,wu2025hunyuanvideo}, we demonstrate that TAGRPO achieves substantial improvements across multiple metrics, establishing a new state-of-the-art for GRPO-based post-training in image-to-video generation.

\section{Related Work}
\label{sec:relatedwork}
\textbf{Image-to-Video Diffusion Models}.
Recent advancements in diffusion-based generative models \cite{ho2020denoising,dhariwal2021diffusion,rombach2022high,flux2024} have extended their capabilities beyond static image synthesis, giving rise to powerful image-to-video (I2V) diffusion frameworks. Unlike text-to-video generation, I2V generation aims to produce temporally coherent motion sequences from one or a few reference images, often guided by corresponding textual prompts.
Early works explored different strategies to achieve this goal. Some studies \cite{voleti2022mcvd,chen2023seine} adopted mask-based approaches to model motion dynamics while preserving static regions in the input image. Others \cite{zhang2023i2vgen,chen2023videocrafter1} leveraged CLIP \cite{radford2021learning} embeddings to extract semantic visual guidance for conditioning the generation process. A separate line of research \cite{blattmann2023stable,zeng2024make} focused on encoding visual embeddings within the VAE latent space to better align appearance and motion consistency across frames.
In recent years, the emergence of efficient training methodologies \cite{lipmanflow,liuflow} and the rapid growth of large-scale video datasets \cite{chen2024panda,wang2024vidprom} and have further accelerated progress, resulting in advanced I2V models \cite{zheng2024open,yang2024cogvideox,shi2024motion,wang2023videocomposer,xing2024dynamicrafter,tian2025extrapolating,guo2024i2v} such as Sora \cite{openai2024sora}, Seedance \cite{gao2025seedance}, Wan \cite{wan2025wan}, Veo \cite{google2024veo}, and HunyuanVideo \cite{kong2024hunyuanvideo}. These systems deliver substantial improvements in visual quality, temporal stability, and motion fidelity.
Despite these remarkable advancements in architecture and training, post-training techniques for image-to-video generation—such as reinforcement learning (RL)—remain underexplored, presenting an important direction for future research.

\textbf{RL for Diffusion Models}.
Research on applying reinforcement learning (RL) techniques to the visual domain has expanded rapidly in recent years. Some approaches \cite{xu2023imagereward,shen2025directly,clark2023directly,prabhudesai2023aligning,prabhudesai2024video} incorporated reward-based optimization, where reward signals are backpropagated through the inference process to refine generative outputs toward desired objectives. Other works \cite{wallace2024diffusion,liu2025videodpo,yang2024using,yuan2024self,zhang2024onlinevpo,furuta2024improving,liang2025aesthetic,du2025reg} extended Direct Preference Optimization (DPO) \cite{rafailov2023direct} to visual generation tasks, aligning model outputs with human preferences.
Building on this progress, recent studies \cite{liu2025flow,xue2025dancegrpo} introduced Group Relative Policy Optimization (GRPO) into the visual domain, leveraging its success in large language models (LLMs) \cite{shao2024deepseekmath} to improve training stability and reward efficiency. Subsequent works further optimized computational cost by refining the rollout procedure \cite{li2025mixgrpo,he2025tempflow,fu2025dynamic,li2025branchgrpo} or by developing feed-forward alternatives that bypass iterative sampling \cite{zheng2025diffusionnft,li2025uniworld,xue2025advantage}.
Despite these promising advances, existing RL-based approaches have predominantly focused on text-conditioned generative tasks, leaving image-to-video generation largely unexplored. This gap highlights an important opportunity for integrating reinforcement learning to enhance the generation quality in I2V diffusion models.

\section{Method}
\label{sec:method}
\subsection{Preliminaries}
\subsubsection{Image-to-Video Diffusion Models}
Image-to-video (I2V) diffusion models extend conventional diffusion-based generators to the spatio-temporal setting, aiming to synthesize temporally coherent motion sequences conditioned on one or more reference images. 
Given a conditional signal $\mathbf{c}$ including an input image and its associated textual prompt, an I2V model produces a corresponding video $\mathbf{x}_0$. The condition image serves as the first frame of the generated video $\mathbf{x}_0$.

Following recent advances in flow-matching frameworks~\cite{lipmanflow,liuflow}, the forward noising process is defined as a linear interpolation between the conditional input and Gaussian noise:
\begin{equation}
\mathbf{x}_t = (1 - t)\mathbf{x}_0 + t\mathbf{x}_1, \quad \mathbf{x}_1 \sim \mathcal{N}(0, \mathbf{I}),
\end{equation}
where $t \in [0, 1]$ represents a time-dependent noise level. 
A neural network $\mathbf{v}_\theta(\mathbf{x}_t, \mathbf{c}, t)$ is trained to estimate the instantaneous velocity that defines the denoising trajectory back toward the clean video sample.

During training, the network receives random clean video samples $\mathbf{x}_0$, noise samples $\mathbf{x}_1$, and conditional signals $\mathbf{c}$. 
The optimization objective corresponds to the flow-matching loss:
\begin{equation}
\mathcal{L}_{\mathrm{I2V}}(\theta) = 
\mathbb{E}_{t, \mathbf{x}_0, \mathbf{x}_1, \mathbf{c}}
\left[\|\mathbf{v}_\theta(\mathbf{x}_t, \mathbf{c}, t) - (\mathbf{x}_1 - \mathbf{x}_0)\|_2^2\right],
\end{equation}
which encourages the network to predict the correct direction of denoising flow while maintaining temporal and structural consistency with the input image.
At inference, a video is generated by numerically integrating the learned ordinary differential equation (ODE):
\begin{equation}
\frac{d\mathbf{x}_t}{dt} = \mathbf{v}_\theta(\mathbf{x}_t, \mathbf{c}, t),
\end{equation}
starting from $\mathbf{x}_1 \sim \mathcal{N}(0, \mathbf{I})$ and solving it backward from $t = 1$ to $t = 0$. 
This process yields a temporally smooth video that preserves the visual identity and semantics of the input image.

\begin{figure*}[t]
  \centering
   \includegraphics[width=1.0\textwidth]{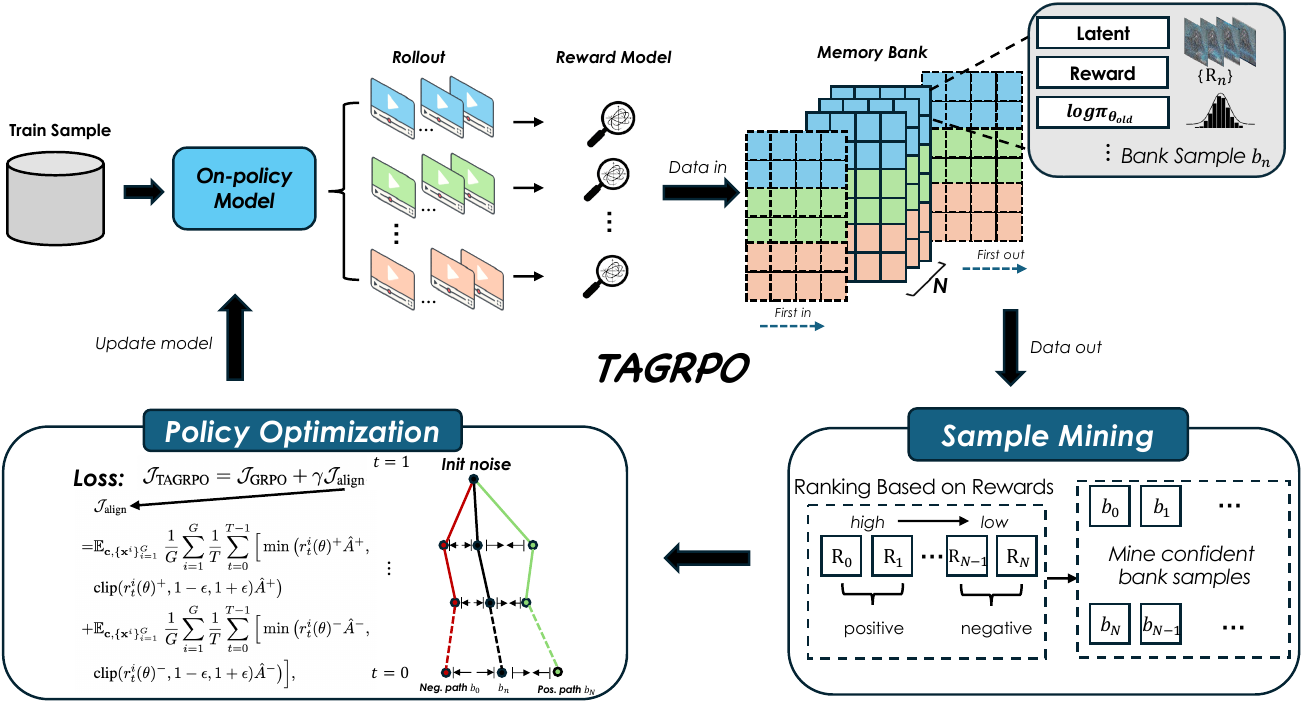}
    \caption{\textbf{Overview of our proposed TAGRPO.} Given a training sample, we generate multiple video samples and evaluate them using a reward model. For each group of samples generated from the same initial noise, we apply both the standard GRPO loss and our trajectory-wise loss $\mathcal{J}_{\text{align}}$ on intermediate latents. $\mathcal{J}_{\text{align}}$ implicitly encourages alignment with high-reward trajectories while maintaining distance from low-reward ones. A memory bank stores historical samples and their rewards, enabling efficient exploitation of diverse past generations without requiring large per-step rollouts. For simplicity, we omit the reference model for computing KL divergence.}
\label{fig:pipeline}
\end{figure*}

\subsubsection{GRPO for Diffusion Models}

Reinforcement learning (RL) aims to optimize a policy that maximizes the expected cumulative reward under a given environment or task condition. 
For diffusion-based generative models, Group Relative Policy Optimization (GRPO) provides an efficient way to align model outputs with human preferences through group-wise reward normalization. 
Instead of optimizing a single trajectory, GRPO jointly considers a batch of samples generated under the same condition, encouraging relative ranking consistency among them.

Given a conditioning signal $\mathbf{c}$ (\textit{e.g.}, a text prompt and a corresponding image), the policy model parameterized by $\theta$ samples a group of $G$ trajectories $\{\left(\mathbf{x}^i_T, \mathbf{x}^i_{T-1}, \ldots, \mathbf{x}^i_0\right)\}_{i=1}^G$.
The optimization objective is defined as follows:
\begin{equation}
\begin{aligned}
&\mathcal{J}_{\text{GRPO}}(\theta)=\mathbb{E}_{\mathbf{c}, \{\mathbf{x}^i\}_{i=1}^G}\;
\frac{1}{G} \sum_{i=1}^{G} \frac{1}{T} \sum_{t=0}^{T-1} \Big[
\min\big(r^i_t(\theta)\hat{A}^i,\, \text{clip}(r^i_t(\theta), 1-\epsilon, 1+\epsilon)\hat{A}^i\big) - \beta D_{\mathrm{KL}}(\pi_\theta || \pi_{\text{ref}})
\Big],
\end{aligned}
\label{eq:grpo_obj}
\end{equation}
where $\hat{A}^i$ denotes the normalized advantage, $\epsilon$ is the clipping coefficient, and $\beta$ controls the strength of the KL regularization term.
The importance ratio between the updated and old policies is computed as:
\begin{equation}
r^i_t(\theta) = 
\frac{\pi_\theta(\mathbf{x}^i_{t-1}|\mathbf{x}^i_t, \mathbf{c})}{
\pi_{\theta_{\text{old}}}(\mathbf{x}^i_{t-1}|\mathbf{x}^i_t, \mathbf{c})}.
\label{eq:ratio}
\end{equation}
For each group of generated samples $\{\mathbf{x}^i_0\}_{i=1}^G$, the group-relative advantage is estimated by normalizing the sample-level rewards:
\begin{equation}
\hat{A}^i = 
\frac{R(\mathbf{x}^i_0, \mathbf{c}) - 
\text{mean}\left(\{R(\mathbf{x}^j_0, \mathbf{c})\}_{j=1}^G\right) }
{\text{std}\left(\{R(\mathbf{x}^j_0, \mathbf{c})\}_{j=1}^G\right)},
\label{eq:advantage}
\end{equation}
where $R(\mathbf{x}^i_0, \mathbf{c})$ represents the reward associated with the generated output $\mathbf{x}^i_0$ conditioned on $\mathbf{c}$.

To stabilize training and encourage better sample diversity, Flow-GRPO \cite{liu2025flow} reformulates the deterministic ordinary differential equation (ODE) of the diffusion process into a stochastic differential equation (SDE) that preserves the same marginal probability distribution at every timestep $t$.
The general form is as follows,
\begin{align}
\mathbf{x}_{t+\Delta t} = \mathbf{x}_t + 
\Big[\mathbf{v}_\theta(\mathbf{x}_t, \mathbf{c}, t)+\frac{\sigma_t^2}{2t}(\mathbf{x}_t + (1-t)\mathbf{v}_\theta(\mathbf{x}_t, \mathbf{c}, t))\Big]\Delta t + \sigma_t \sqrt{\Delta t} \boldsymbol{\epsilon},
\label{eq:sde}
\end{align}
where $\boldsymbol{\epsilon} \sim \mathcal{N}(0, \mathbf{I})$ introduces stochasticity, and $\sigma_t$ denotes the noise scale.
The KL divergence between the current policy $\pi_\theta$ and a reference policy $\pi_{\text{ref}}$ admits the following closed-form approximation:
\begin{align}
D_{\mathrm{KL}}(\pi_\theta || \pi_{\text{ref}})= \frac{\Delta t}{2}
\left(
\frac{\sigma_t(1-t)}{2t} + \frac{1}{\sigma_t}
\right)^2
\|\mathbf{v}_\theta(\mathbf{x}_t, \mathbf{c}, t) - \mathbf{v}_{\text{ref}}(\mathbf{x}_t, \mathbf{c}, t)\|_2^2.
\label{eq:kl}
\end{align}
Together, these formulations allow GRPO to align diffusion-based video or image generators with reward functions while maintaining stable and consistent optimization dynamics.

\subsection{TAGRPO}

Although previous GRPO-based approaches have achieved success in the visual domain, they have predominantly focused on text-conditioned generative models, overlooking the image-to-video (I2V) diffusion setting. To the best of our knowledge, DanceGRPO \cite{xue2025dancegrpo} is the only method that has been implemented for an I2V model, \textit{i.e.}, SkyReels-I2V \cite{SkyReelsV1}, which represents a relatively weak baseline. Crucially, our experiments reveal that directly applying DanceGRPO to state-of-the-art I2V architectures—such as Wan 2.2 \cite{wan2025wan} and HunyuanVideo 1.5 \cite{wu2025hunyuanvideo}—fails to yield meaningful improvements. These findings indicate that post-training for image-to-video models remains an open challenge, necessitating specialized optimization strategies.

To address this, we present TAGRPO, an effective framework for post-training I2V models based on the idea of \textbf{T}rajectory \textbf{A}lignment, as shown in Figure \ref{fig:pipeline}. 
Our main motivation is that exploiting these inter-sample relationships can significantly boost optimization. 
Specifically, we identify the video latents with the highest ($\mathbf{x}^{+}_t$) and lowest ($\mathbf{x}^{-}_t$) rewards within a group and treat them as global positive and negative anchors of the group. Consequently, every latent $\mathbf{x}^i_t$ in the group is optimized to align its trajectory with the best sample while diverging from the worst. Mathematically, we introduce a trajectory alignment loss, $\mathcal{J}_{\text{align}}$, defined as:
\begin{align}
\mathcal{J}_{\text{align}} = &\mathbb{E}_{\mathbf{c}, \{\mathbf{x}^i\}_{i=1}^G}\;
\frac{1}{G} \sum_{i=1}^{G} \frac{1}{T} \sum_{t=0}^{T-1} \Big[
\min\big(r^i_t(\theta)^{+}\hat{A}^{+},\, \text{clip}(r^i_t(\theta)^{+}, 1-\epsilon, 1+\epsilon)\hat{A}^{+}\big) \\
+ & \mathbb{E}_{\mathbf{c}, \{\mathbf{x}^i\}_{i=1}^G}\;
\frac{1}{G} \sum_{i=1}^{G} \frac{1}{T} \sum_{t=0}^{T-1} \Big[
\min\big(r^i_t(\theta)^{-}\hat{A}^{-},\, \text{clip}(r^i_t(\theta)^{-}, 1-\epsilon, 1+\epsilon)\hat{A}^{-}\big)
\Big],
\label{eq:align_loss}
\end{align}
where $\hat{A}^{+}$ and $\hat{A}^{-}$ denote the normalized advantage of the most positive and negative generated videos, respectively. The importance ratios $r^i_t(\theta)^{+}$ and $r^i_t(\theta)^{-}$ measure the likelihood of sample $i$ following the positive or negative trajectory of the group:
\begin{equation}
r^i_t(\theta)^{+} = 
\frac{\pi_\theta(\mathbf{x}^{+}_{t-1}|\mathbf{x}^i_t, \mathbf{c})}{
\pi_{\theta_{\text{old}}}(\mathbf{x}^{+}_{t-1}|\mathbf{x}^i_t, \mathbf{c})}.
\label{eq:ratio_pos}
\end{equation}
\begin{equation}
r^i_t(\theta)^{-} = 
\frac{\pi_\theta(\mathbf{x}^{-}_{t-1}|\mathbf{x}^i_t, \mathbf{c})}{
\pi_{\theta_{\text{old}}}(\mathbf{x}^{-}_{t-1}|\mathbf{x}^i_t, \mathbf{c})}.
\label{eq:ratio_neg}
\end{equation}
This formulation implicitly encourages all generated samples in the same group to mimic the transitions of the most positive trajectory and avoid those of the most negative one, providing effective directional guidance based on inter-sample relationships. The final objective function is:
\begin{align}
\mathcal{J}_{\text{TAGRPO}} = \mathcal{J}_{\text{GRPO}} + \gamma \mathcal{J}_{\text{align}}.
\label{eq:total_loss}
\end{align}

\begin{figure*}[h!]
  \centering
   \includegraphics[width=1.0\textwidth]{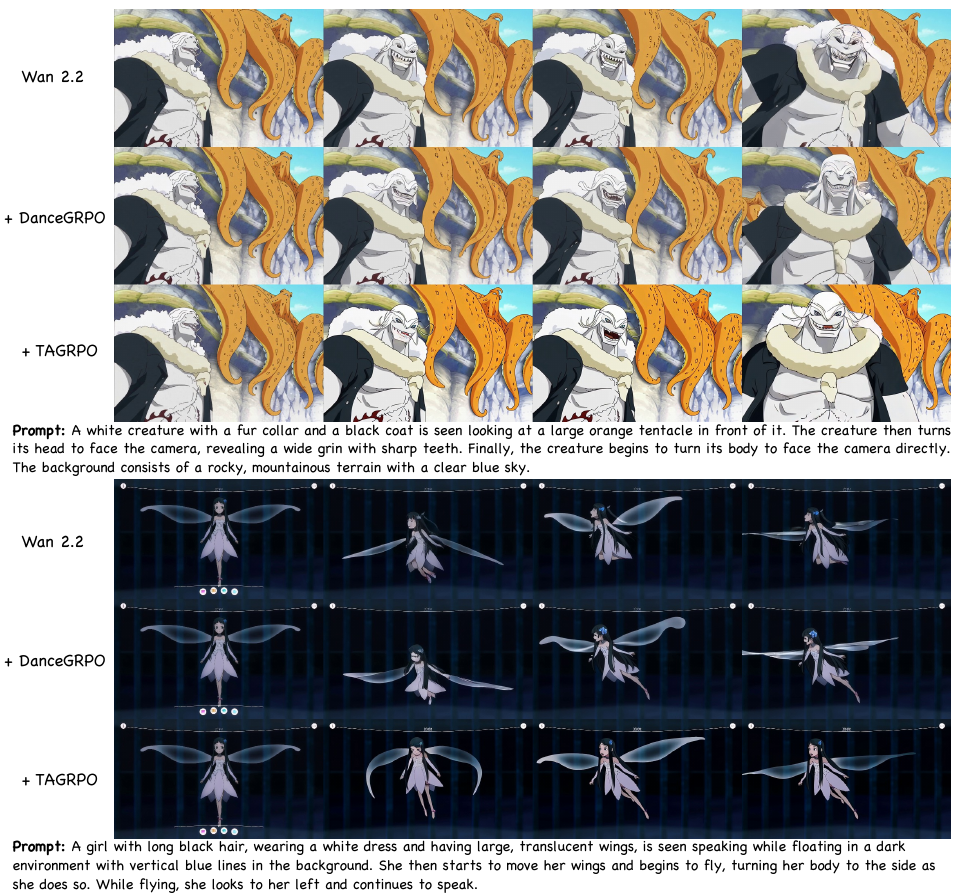}
   \caption{Qualitative comparison among TAGRPO, DanceGRPO and the base model Wan 2.2. Models trained with TAGRPO demonstrate superior visual quality with improved aesthetics, reduced distortion artifacts and better motion realism in animation scenes.}
   \label{fig:cmp_qualitative1}
\end{figure*}

To maximize the effectiveness of $\mathcal{J}_{\text{align}}$, sufficient rollout videos with diverse rewards are essential. However, generating these videos per step incurs significant computational costs and substantially slows the training process.
Inspired by contrastive learning principles \cite{he2020momentum}, we propose maintaining a memory bank for TAGRPO that stores previously generated video latents and their corresponding reward signals from past iterations. This approach enables us to accumulate a diverse collection of generated videos while keeping the per-step generation count low, thereby reducing computational overhead.
Furthermore, this memory mechanism helps prevent the model from diverging significantly from its original parameters by leveraging previously generated samples, thus providing optimization stability.

\section{Experiments}
\label{sec:experiments}
\subsection{Implementation Details}
We applied our method to advanced image-to-video models, Wan 2.2 \cite{wan2025wan} and HunyuanVideo 1.5 \cite{wu2025hunyuanvideo}.
As demonstrated in previous studies \cite{he2025tempflow,li2025mixgrpo}, higher timesteps exert a greater influence on the quality of generated videos.
Consequently, for Wan 2.2, we propose to optimize the high-noise model of Wan 2.2 while keeping its low-noise counterpart unchanged.
For HunyuanVideo 1.5, we also optimize the early timestep values, which are greater than $900$ in experiments.

Regarding reward models, given the scarcity of effective open-source reward models designed specifically for image-to-video generative models, we leveraged the image reward model, HPSv3 \cite{ma2025hpsv3} and the Q-Save evaluation model \cite{wu2025q} in our experiments.
For HPSv3, we uniformly sampled two frames per second from each generated video and computed the average reward across these frames as the overall video reward.
For Q-Save, we used the combination of Visual Quality (VQ), Dynamic Quality (DQ) and Image Alignment (IA) as the overall reward for generated videos.

Following previous studies \cite{xue2025dancegrpo,he2025tempflow,li2025mixgrpo}, we focus on samples generated from the same initial noise to control variations in the inference process.
We set the group size $G=8$, with hyperparameters $\gamma=1$. 
To speed up the training process, we set the training resolution of generated videos as $320$p with $53$ frames in total.
The inference step number was set $16$ and the classifier free guidance was set $3.5$.
For $\mathcal{J}_{\text{align}}$, $x_t^+$ was set as the latent representation of the video achieving the highest reward, while $x_t^-$ corresponded to the latent of the video with the lowest reward within that group.

For the memory bank, we implemented a first-in-first-out (FIFO) strategy to continuously refresh stored samples, thereby maintaining both relevance and diversity throughout training. 
For training data, we used an internal dataset containing approximately 10K image-text pairs, featuring diverse scenes and image styles. The dataset encompasses a wide range of visual content, including natural landscapes, urban environments, portraits, anime, and abstract compositions, with corresponding text descriptions that vary in length and complexity to ensure robust learning.
Moreover, to evaluate the effectiveness of our proposed method, we then subsampled an evaluation set from our training data, dubbed \textit{TAGRPO-Bench}, containing $200$ challenging image-text pairs for the task of image-to-video generation.

\subsection{Qualitative Comparisons}

\begin{figure*}[t]
  \centering
   \includegraphics[width=1.0\textwidth]{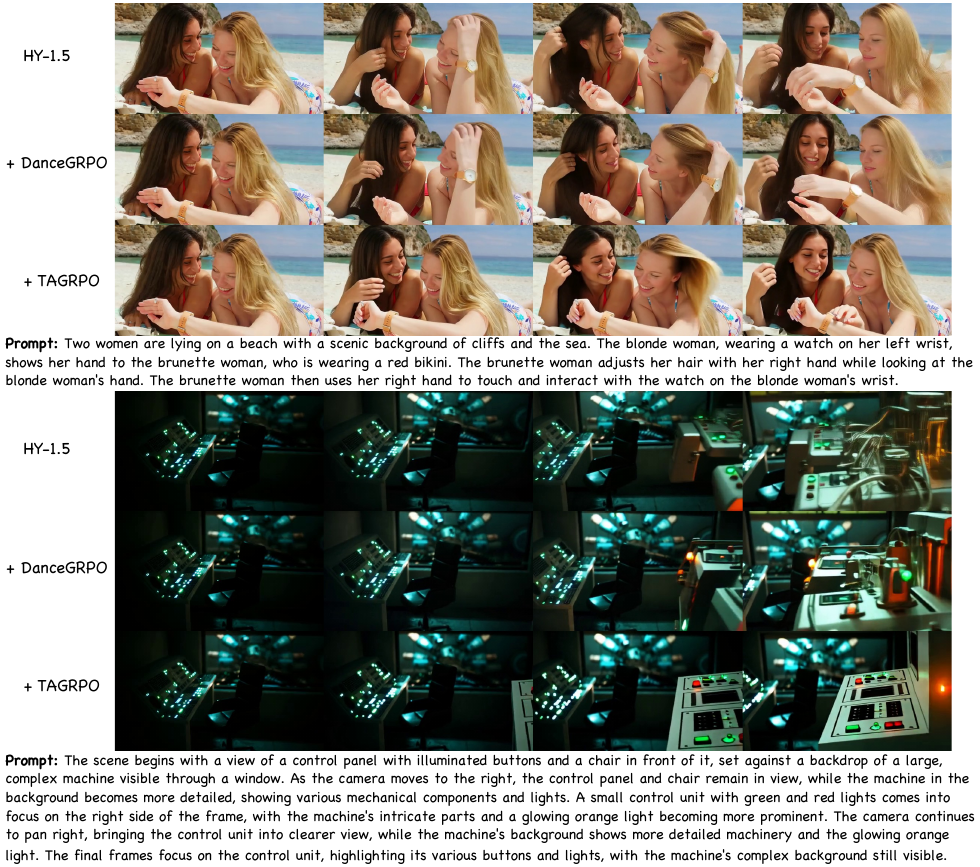}
   \caption{Qualitative comparison among TAGRPO, DanceGRPO and the base model HunyuanVideo 1.5 (HY-1.5). Models trained with TAGRPO exhibit superior generation fidelity, characterized by sharper structural details and significantly fewer temporal artifacts.}
   \label{fig:cmp_qualitative2}
\end{figure*}

We present visual comparisons on our TAGRPO-Bench to evaluate the perceptual quality and prompt adherence of generated videos across different backbones.
Figures~\ref{fig:cmp_qualitative1} and \ref{fig:cmp_qualitative2} present qualitative comparisons on Wan 2.2 and HunyuanVideo-1.5, respectively. In Figure~\ref{fig:cmp_qualitative1}, TAGRPO demonstrates superior motion control: it executed the creature's head turn with a coherent ``wide grin" and maintains the fairy's anatomical correctness, whereas baselines suffered from significant facial distortions. Figure~\ref{fig:cmp_qualitative2} highlights fidelity and stability; TAGRPO preserved sharp details in the blonde hair (top) and maintained rigid geometric consistency during the sci-fi camera pan, avoiding the structural warping and texture drift observed in other models. These results confirm that our trajectory alignment mechanism effectively pruned generation paths leading to visual artifacts and instability. More visual comparisons are provided \href{https://tagrpo.github.io/}{here}.

\subsection{Quantitative Comparisons}
In this section, we performed quantitative comparisons to evaluate TAGRPO against both the base model and DanceGRPO on the TAGRPO-Bench, utilizing the HunyuanVideo-1.5 (HY-1.5) and Wan 2.2 backbones across both 320p and 720p resolutions. 
As summarized in Table \ref{tab:HY_comparison} and Table \ref{tab:wan22_comparison}, our method consistently achieved the best performance across both backbones, demonstrating the effectiveness of our trajectory alignment strategy. 
Notably, although our models were trained exclusively under the 320p setting, they still achieved significant improvements at 720p, highlighting the strong generalization capability of our approach.

\begin{table}[t]
\centering

\begin{minipage}[t]{0.48\textwidth}
    \centering
    \caption{Quantitative comparison on the HunyuanVideo 1.5 (HY-1.5) baseline. We evaluated performance using Q-Save and HPSv3 metrics across 320p and 720p resolutions. TAGRPO consistently outperformed both the base model and DanceGRPO.}
    \label{tab:HY_comparison}
    {
    \linespread{1.3}
    \setlength{\tabcolsep}{4pt}
    \small
    \begin{tabular}{llccc}
    \toprule
    \textbf{Metric} & \textbf{Res.} & \textbf{HY-1.5} & \textbf{+DanceGRPO} & \textbf{+TAGRPO} \\
    \midrule
    \multirow{2}{*}{Q-Save}& {320p} &8.01 &  8.01  &\textbf{8.05} \\
    \cmidrule(lr){2-5}
    & 720p   &10.02  & 10.02 &\textbf{10.05}  \\
    \midrule
    \multirow{2}{*}{HPSv3}& {320p} &2.00 &1.84  & \textbf{2.41} \\
    \cmidrule(lr){2-5}
    & 720p   & 4.42 & 4.33 & \textbf{4.58} \\
    \bottomrule
    \end{tabular}
    }
\end{minipage}%
\hfill 
\begin{minipage}[t]{0.48\textwidth}
    \centering
    \caption{Quantitative comparison on the Wan 2.2 baseline. We reported Q-Save and HPSv3 scores at 320p and 720p resolutions. TAGRPO demonstrated robust improvements over the baseline and DanceGRPO, achieving the highest scores in all settings.}
    \label{tab:wan22_comparison}
    {
    \linespread{1.3}
    \setlength{\tabcolsep}{4pt}
    \small
    \begin{tabular}{llccc}
    \toprule
    \textbf{Metric} & \textbf{Res.} & \textbf{Wan 2.2} & \textbf{+DanceGRPO} & \textbf{+TAGRPO} \\
    \midrule
    \multirow{2}{*}{Q-Save}& {320p}&8.73 &8.75  &\textbf{8.81}  \\
    \cmidrule(lr){2-5}
    & 720p   & 10.13 & 10.17 &\textbf{10.26} \\
    \midrule
    \multirow{2}{*}{HPSv3}& {320p} &3.63&3.70&\textbf{4.29}   \\
    \cmidrule(lr){2-5}
    & 720p   & 4.34 &4.40 &\textbf{5.03}  \\
    \bottomrule
    \end{tabular}
    }
\end{minipage}

\end{table}

\subsection{Ablation Studies}
In this section, we performed ablation studies on the effectiveness of our proposed methods.
\subsubsection{Components Effectiveness}
\label{subsubsection: Components}
\begin{figure}[t]
  \centering
  
  \begin{minipage}[t]{0.48\textwidth}
    \centering
    \includegraphics[width=\textwidth]{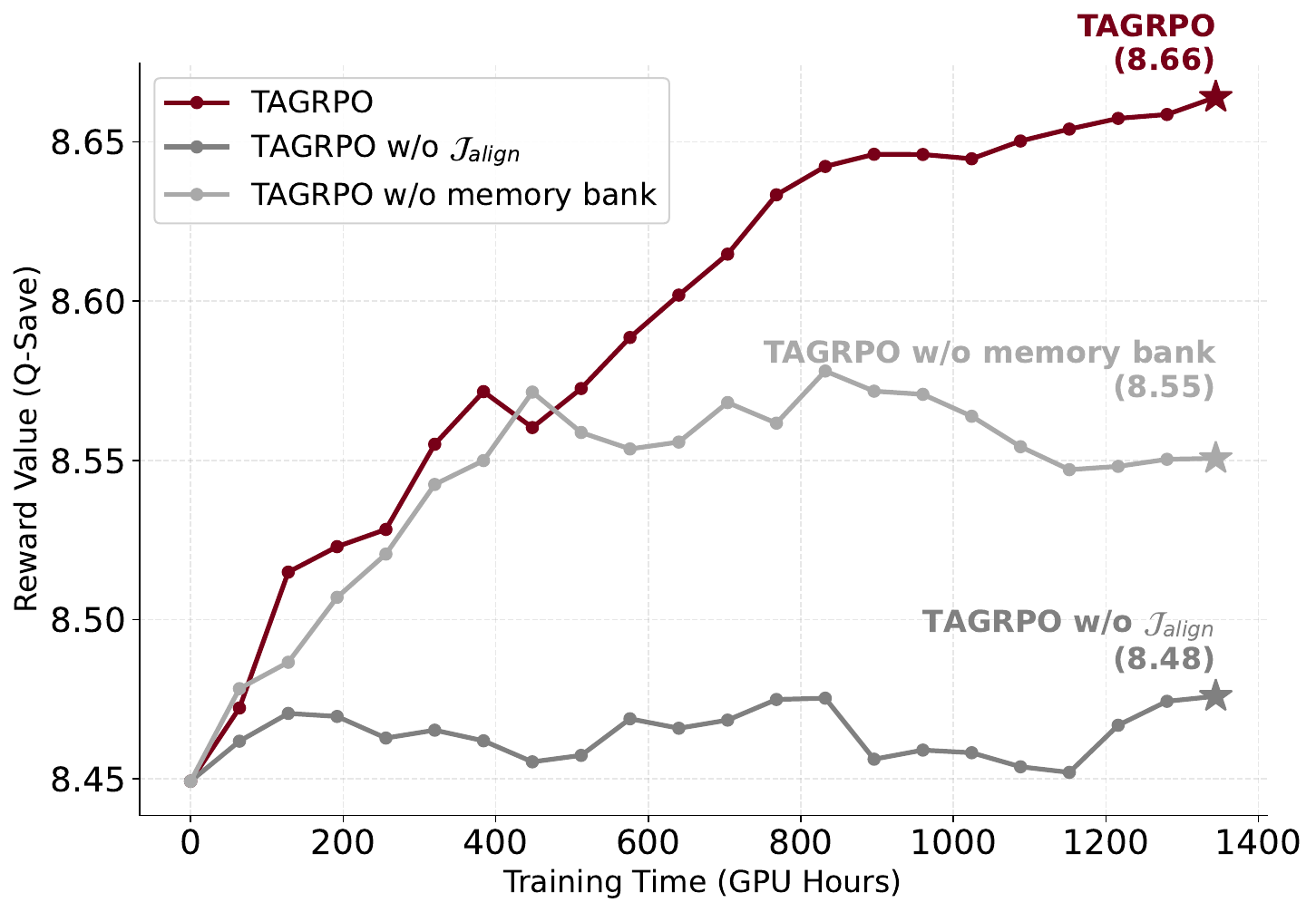}
    \caption{Ablation study on the contributions of $\mathcal{J}_{\text{align}}$ loss and memory bank mechanism. TAGRPO achieves the highest reward improvement, while removing either component results in slower convergence and lower final performance.}
    \label{fig:ablation1}
  \end{minipage}%
  \hfill 
  \begin{minipage}[t]{0.48\textwidth}
    \centering
    \includegraphics[width=\textwidth]{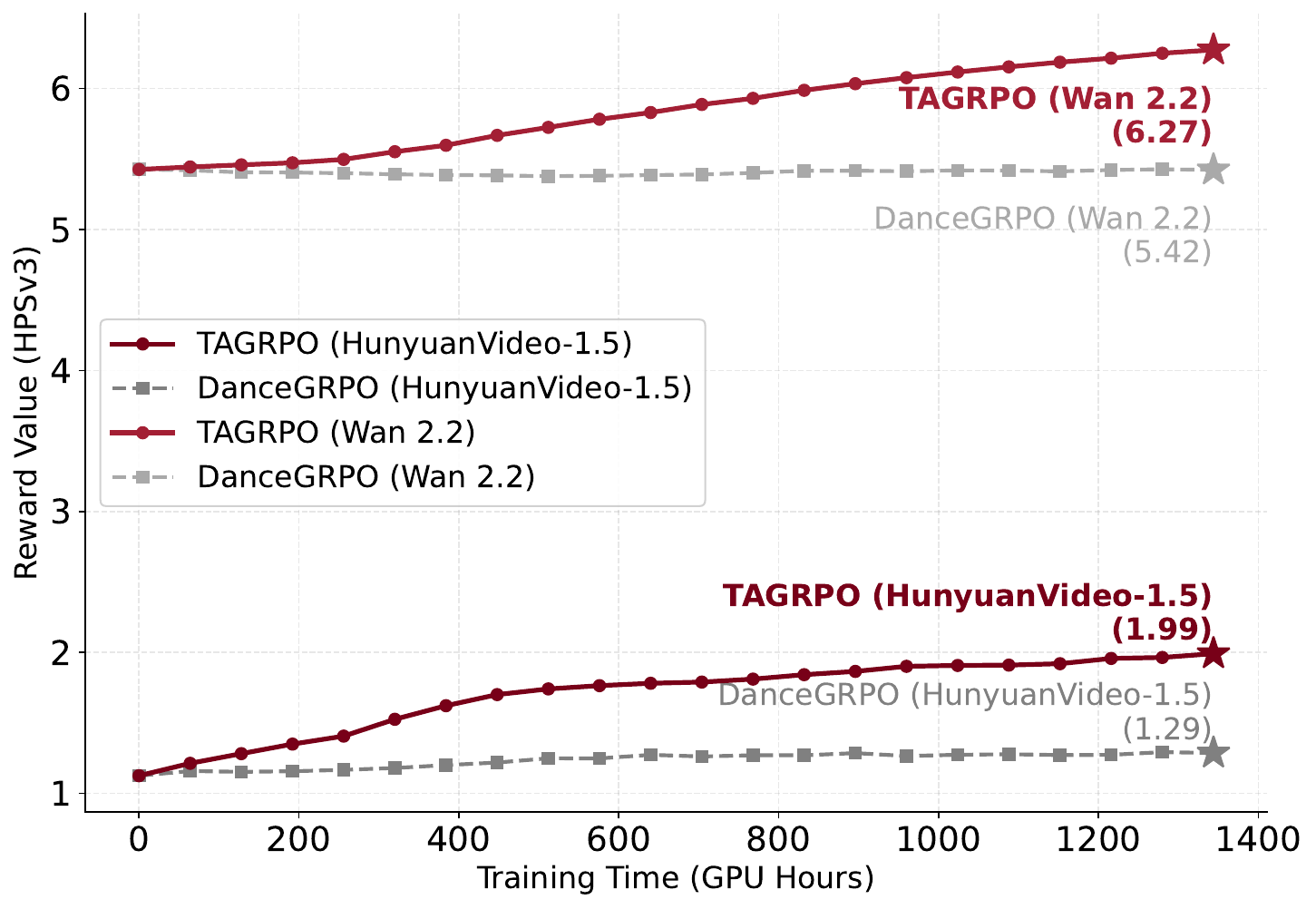}
    \caption{Generalization to other settings. We applied our method to state-of-the-art text-conditioned AIGC models. Our method demonstrated faster convergence and achieved higher final rewards, showing the potential of our approach to other AIGC tasks.}
    \label{fig:ablation3}
  \end{minipage}
  
\end{figure}

In this section, we conducted an ablation study to evaluate the contributions of the $\mathcal{J}_{\text{align}}$ loss and memory bank mechanism in our proposed TAGRPO.
As illustrated in Figure \ref{fig:ablation1}, we conducted three experiments comparing: (1) TAGRPO, (2) TAGRPO without the memory bank, and (3) TAGRPO without $\mathcal{J}_{\text{align}}$, using the  combination of Visual Quality (VQ), Dynamic Quality (DQ) and Image Alignment (IA) in Q-Save \cite{wu2025q} as the reward metrics.
The reported values in the figure were averaged over the evaluation dataset, which was a small subset of our internal training data.
The results demonstrate that TAGRPO achieved the greatest reward improvement, confirming the necessity of each component.
Specifically, removing either the memory bank or $\mathcal{J}_{\text{align}}$ resulted in slower convergence and lower reward values, indicating that both components play roles in effective policy optimization.
This validates our hypothesis that combining trajectory-wise supervision with diverse historical samples leads to more robust and efficient fine-tuning.

\subsubsection{Generalization to Other Settings} 
To demonstrate the generalization capability of our proposed method, we conducted new experiments by extending our approach to text to video tasks, including Wan2.2-T2V-A14B~\cite{wan2025wan} and HunyuanVideo-1.5-720P-T2V~\cite{wu2025hunyuanvideo}.
We still used HPSv3 \cite{ma2025hpsv3} as the reward model.
To maintain the content similarity among rollout videos in each group, we also used the same initial noise for video generations.
As shown in Figure~\ref{fig:ablation3}, we compared our method against DanceGRPO~\cite{xue2025dancegrpo} when applied to this alternative setting.
The results demonstrate that our method still achieved faster convergence and higher final reward scores, indicating the potential our approach to other AIGC settings.

\subsubsection{Analysis on the Memory Bank Design} 
\label{section:memorybank_design}
\begin{figure}[t]
  \centering
  
  \begin{minipage}[t]{0.32\textwidth}
    \centering
    \includegraphics[width=\textwidth]{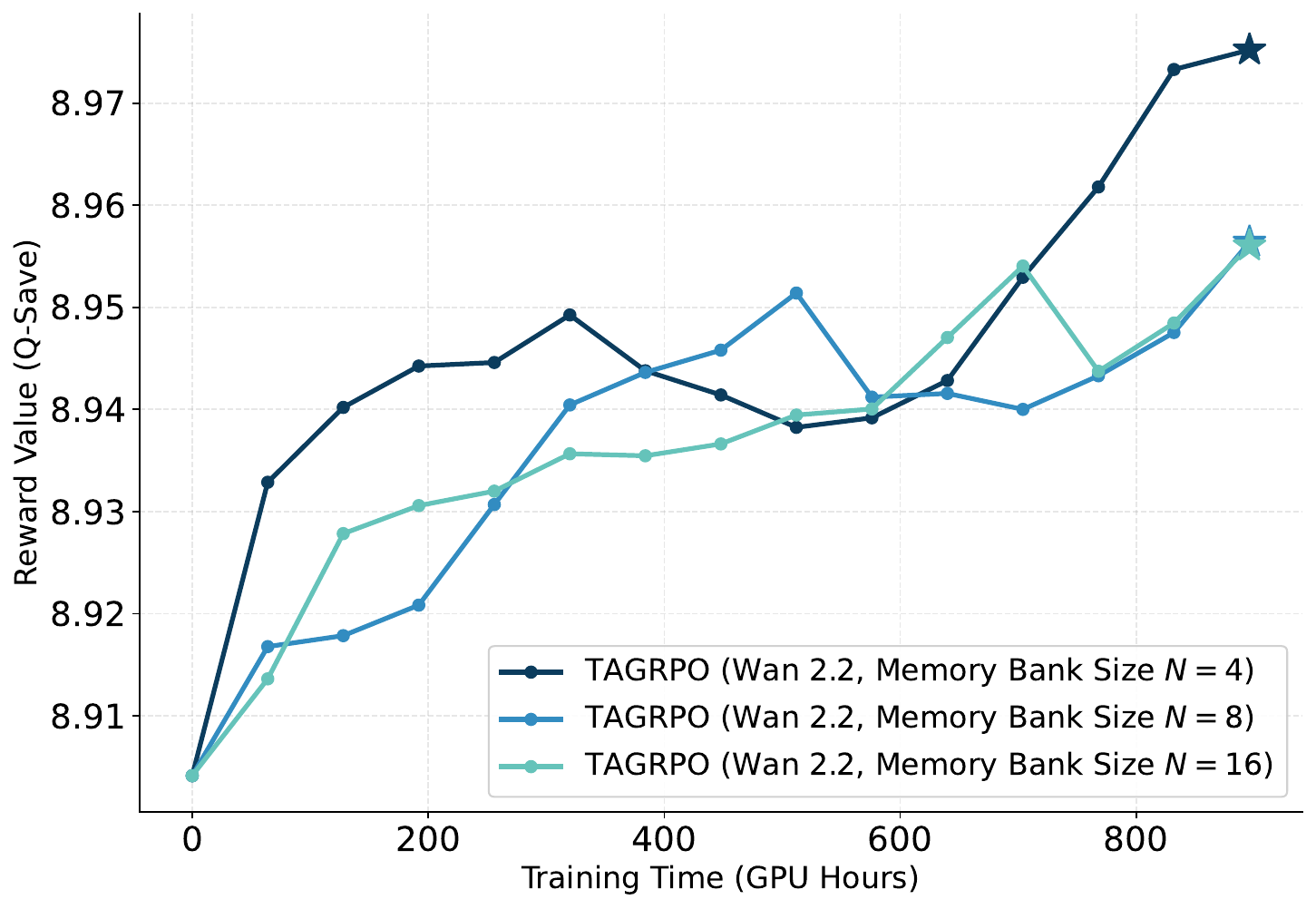}
    \caption{Ablation study on the memory bank size ($N$). The reward curves demonstrate a consistent upward trend across all evaluated configurations, highlighting the robustness of TAGRPO.}
    \label{fig:ablationMembank}
  \end{minipage}%
  \hfill 
  \begin{minipage}[t]{0.32\textwidth}
    \centering
    \includegraphics[width=\textwidth]{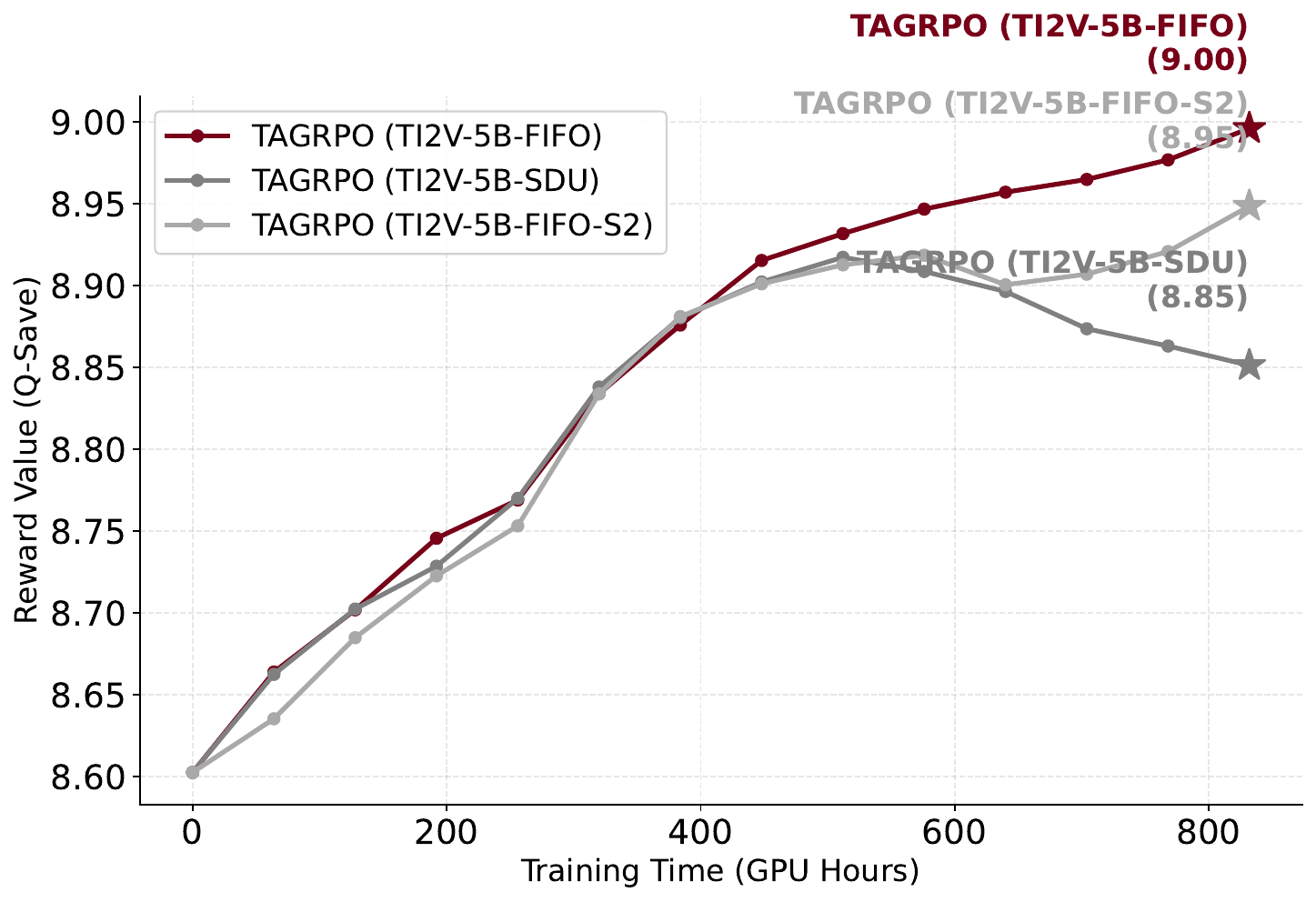}
    \caption{Ablation on the memory bank update strategies (FIFO vs. SDU) and frequency (FIFO vs. FIFO-S2). The default FIFO strategy provided the most stable and highest reward convergence.}
    \label{fig:ablationMembank2}
  \end{minipage}%
  \hfill 
  \begin{minipage}[t]{0.32\textwidth}
    \centering
    \includegraphics[width=\textwidth]{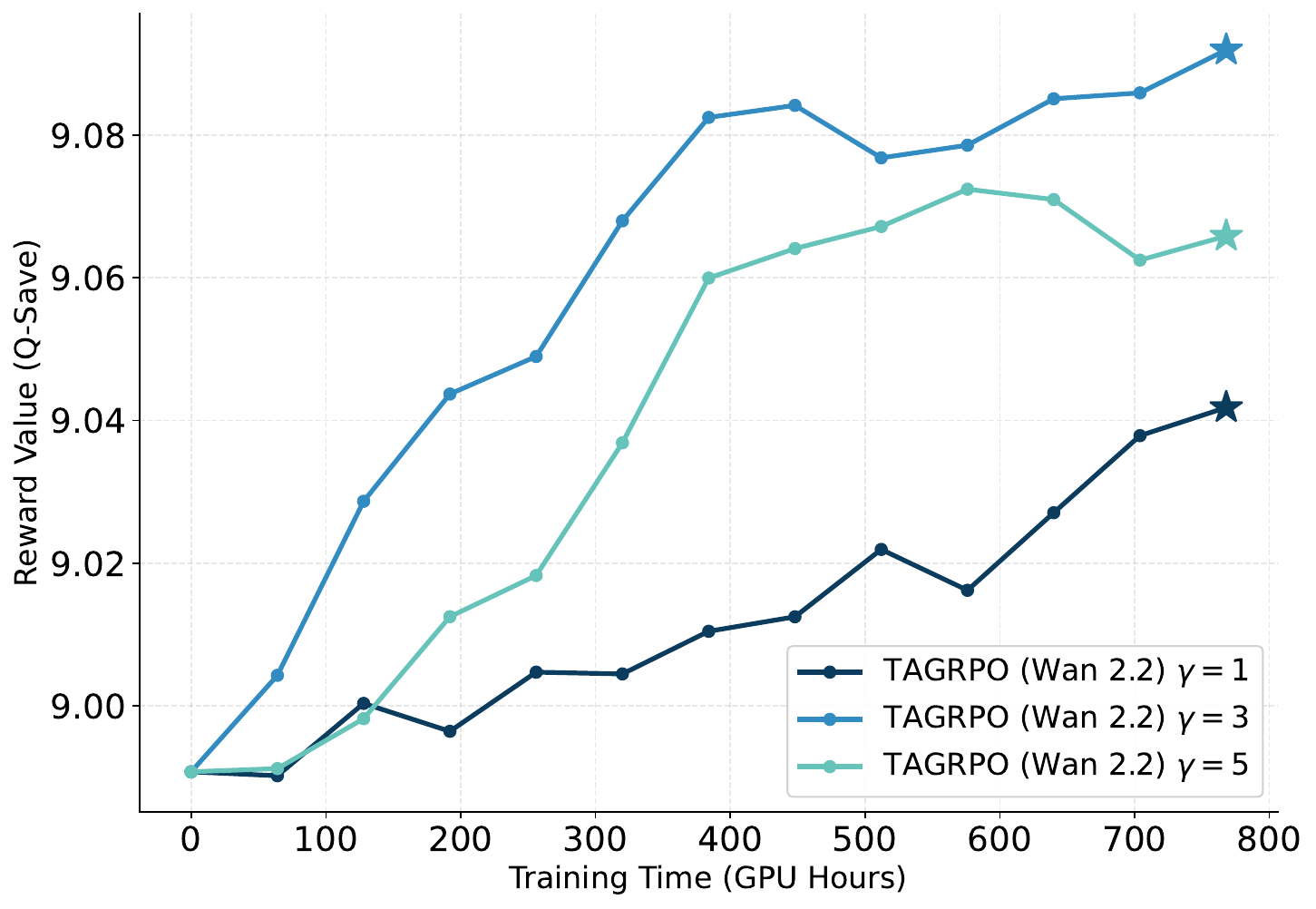}
    \caption{Sensitivity analysis of $\gamma$. TAGRPO demonstrated stable reward convergence across various $\gamma$ configurations.}
    \label{fig:ablationgamma}
  \end{minipage}
  
\end{figure}

To evaluate the impact of the memory bank's capacity ($N$), we conducted experiments on a training subset using Wan 2.2 as the base model. Specifically, we compared configurations where $N \in \{4, 8, 16\}$\footnote{Note that each group of $G$ trajectories acts as a single memory unit here. Thus, a capacity of $N$ means the memory bank stores up to $N \times G$ trajectories, from which the highest ($\mathbf{x}^{+}_t$) and lowest ($\mathbf{x}^{-}_t$) reward latents are identified.}.
As illustrated in Figure \ref{fig:ablationMembank}, the reward curves for all configurations exhibited a consistent upward trend, demonstrating that TAGRPO remained robust across various memory bank sizes.

Besides, to analyze the memory bank's update strategies and frequency, we also compared our default First-In-First-Out (FIFO) strategy, which updates the memory bank at every step, against two variants: 1) FIFO-S2 (to evaluate update frequency), which is identical to FIFO but updates occur every 2 training steps; 2) SDU (Score-Driven Update, to evaluate update strategy), which sorts all trajectories in the memory bank by their rewards, intentionally discarding mid-range samples to retain only the highest and lowest reward trajectories when the memory bank is full.
In experiments, we chose the Wan2.2-TI2V-5B \cite{wan2025wan} model as the base architecture and conducted these ablations on a subset of our training data. As shown in Figure \ref{fig:ablationMembank2}, our results yield three key findings: 1) The default FIFO strategy consistently improved the reward, further demonstrating that our method generalized effectively to other base models, like Wan2.2-TI2V-5B.
2) FIFO-S2 converged slower and achieved a lower final reward compared to FIFO. Less frequent updates forced the memory bank to retain ``stale" off-policy trajectories longer. In standard GRPO, utilizing these outdated samples degraded the accuracy of advantage estimation; thus, step-by-step updates were essential to maintain the near on-policy nature required for optimal refinement.
3) SDU exhibited instability on reward improvements. By hoarding only extreme trajectories, SDU prevented the memory bank from refreshing regularly, causing these retained samples to become outdated (stale). 
Furthermore, it created an unnatural bimodal reward distribution. This dual effect distorted the mean and variance used for GRPO advantage normalization, rendering the gradients of newly generated on-policy samples uninformative and causing late-stage training collapse.

\subsubsection{Sensitivity Analysis of $\gamma$}
\label{section:Sensitivity_gamma}
The hyperparameter $\gamma$ plays a pivotal role in balancing the standard policy optimization with our proposed trajectory alignment objective. To systematically evaluate its impact on the training dynamics, we tracked the reward curves across a diverse set of $\gamma$ configurations. As illustrated in Figure \ref{fig:ablationgamma}, the optimization process exhibited a remarkably consistent upward trend and stable convergence across all tested values. This empirical evidence demonstrates that TAGRPO achieved highly robust performance across a wide range of $\gamma$ settings, effectively eliminating the need for exhaustive, model-specific hyperparameter tuning.

\section{Conclusion}
\label{sec:conclusion}
In this paper, we have presented TAGRPO, a novel RL-based post-training framework for image-to-video generation. By introducing a trajectory-wise alignment loss and a memory bank mechanism, TAGRPO has effectively exploited the relative relationships among generated and historical samples, achieving significant improvements over DanceGRPO. Our core insight is that explicitly aligning intermediate denoising trajectories based on reward rankings provides more informative optimization signals than treating samples independently. We believe TAGRPO establishes a new paradigm for efficient alignment in video generation, opening promising avenues for other image-conditioned multimodal tasks.

\section*{Impact Statement}
This paper presents work whose goal is to advance the field of image-to-video generative models, benefiting creative industries, animation, and content creation. While high-fidelity video generation carries inherent risks of misuse, such as deepfakes and visual misinformation, our reward-based framework (TAGRPO) also offers a pathway for mitigation. By integrating safety-oriented reward models, our approach can be actively deployed to align video generators toward safer and more responsible outputs.


\bibliography{example_paper}

@inproceedings{lipmanflow,
  title={Flow Matching for Generative Modeling},
  author={Lipman, Yaron and Chen, Ricky TQ and Ben-Hamu, Heli and Nickel, Maximilian and Le, Matthew},
  booktitle={The Eleventh International Conference on Learning Representations}
}

@inproceedings{liuflow,
  title={Flow Straight and Fast: Learning to Generate and Transfer Data with Rectified Flow},
  author={Liu, Xingchao and Gong, Chengyue and others},
  booktitle={The Eleventh International Conference on Learning Representations}
}

@inproceedings{esser2024scaling,
  title={Scaling rectified flow transformers for high-resolution image synthesis},
  author={Esser, Patrick and Kulal, Sumith and Blattmann, Andreas and Entezari, Rahim and M{\"u}ller, Jonas and Saini, Harry and Levi, Yam and Lorenz, Dominik and Sauer, Axel and Boesel, Frederic and others},
  booktitle={Forty-first international conference on machine learning},
  year={2024}
}

@inproceedings{rombach2022high,
  title={High-resolution image synthesis with latent diffusion models},
  author={Rombach, Robin and Blattmann, Andreas and Lorenz, Dominik and Esser, Patrick and Ommer, Bj{\"o}rn},
  booktitle={Proceedings of the IEEE/CVF conference on computer vision and pattern recognition},
  pages={10684--10695},
  year={2022}
}

@misc{labs2025flux1kontextflowmatching,
      title={FLUX.1 Kontext: Flow Matching for In-Context Image Generation and Editing in Latent Space},
      author={Black Forest Labs and Stephen Batifol and Andreas Blattmann and Frederic Boesel and Saksham Consul and Cyril Diagne and Tim Dockhorn and Jack English and Zion English and Patrick Esser and Sumith Kulal and Kyle Lacey and Yam Levi and Cheng Li and Dominik Lorenz and Jonas Müller and Dustin Podell and Robin Rombach and Harry Saini and Axel Sauer and Luke Smith},
      year={2025},
      eprint={2506.15742},
      archivePrefix={arXiv},
      primaryClass={cs.GR},
      url={https://arxiv.org/abs/2506.15742},
}

@misc{flux2024,
    author={Black Forest Labs},
    title={FLUX},
    year={2024},
}

@article{ho2020denoising,
  title={Denoising diffusion probabilistic models},
  author={Ho, Jonathan and Jain, Ajay and Abbeel, Pieter},
  journal={Advances in neural information processing systems},
  volume={33},
  pages={6840--6851},
  year={2020}
}

@article{dhariwal2021diffusion,
  title={Diffusion models beat gans on image synthesis},
  author={Dhariwal, Prafulla and Nichol, Alexander},
  journal={Advances in neural information processing systems},
  volume={34},
  pages={8780--8794},
  year={2021}
}

@inproceedings{songscore,
  title={Score-Based Generative Modeling through Stochastic Differential Equations},
  author={Song, Yang and Sohl-Dickstein, Jascha and Kingma, Diederik P and Kumar, Abhishek and Ermon, Stefano and Poole, Ben},
  booktitle={International Conference on Learning Representations}
}

@inproceedings{peebles2023scalable,
  title={Scalable diffusion models with transformers},
  author={Peebles, William and Xie, Saining},
  booktitle={Proceedings of the IEEE/CVF international conference on computer vision},
  pages={4195--4205},
  year={2023}
}

@article{blattmann2023stable,
  title={Stable video diffusion: Scaling latent video diffusion models to large datasets},
  author={Blattmann, Andreas and Dockhorn, Tim and Kulal, Sumith and Mendelevitch, Daniel and Kilian, Maciej and Lorenz, Dominik and Levi, Yam and English, Zion and Voleti, Vikram and Letts, Adam and others},
  journal={arXiv preprint arXiv:2311.15127},
  year={2023}
}

@article{ho2022video,
  title={Video diffusion models},
  author={Ho, Jonathan and Salimans, Tim and Gritsenko, Alexey and Chan, William and Norouzi, Mohammad and Fleet, David J},
  journal={Advances in neural information processing systems},
  volume={35},
  pages={8633--8646},
  year={2022}
}

@article{yang2024cogvideox,
  title={Cogvideox: Text-to-video diffusion models with an expert transformer},
  author={Yang, Zhuoyi and Teng, Jiayan and Zheng, Wendi and Ding, Ming and Huang, Shiyu and Xu, Jiazheng and Yang, Yuanming and Hong, Wenyi and Zhang, Xiaohan and Feng, Guanyu and others},
  journal={arXiv preprint arXiv:2408.06072},
  year={2024}
}

@article{kong2024hunyuanvideo,
  title={Hunyuanvideo: A systematic framework for large video generative models},
  author={Kong, Weijie and Tian, Qi and Zhang, Zijian and Min, Rox and Dai, Zuozhuo and Zhou, Jin and Xiong, Jiangfeng and Li, Xin and Wu, Bo and Zhang, Jianwei and others},
  journal={arXiv preprint arXiv:2412.03603},
  year={2024}
}

@article{wan2025wan,
  title={Wan: Open and advanced large-scale video generative models},
  author={Wan, Team and Wang, Ang and Ai, Baole and Wen, Bin and Mao, Chaojie and Xie, Chen-Wei and Chen, Di and Yu, Feiwu and Zhao, Haiming and Yang, Jianxiao and others},
  journal={arXiv preprint arXiv:2503.20314},
  year={2025}
}

@article{zheng2024open,
  title={Open-sora: Democratizing efficient video production for all},
  author={Zheng, Zangwei and Peng, Xiangyu and Yang, Tianji and Shen, Chenhui and Li, Shenggui and Liu, Hongxin and Zhou, Yukun and Li, Tianyi and You, Yang},
  journal={arXiv preprint arXiv:2412.20404},
  year={2024}
}

@article{lin2024open,
  title={Open-sora plan: Open-source large video generation model},
  author={Lin, Bin and Ge, Yunyang and Cheng, Xinhua and Li, Zongjian and Zhu, Bin and Wang, Shaodong and He, Xianyi and Ye, Yang and Yuan, Shenghai and Chen, Liuhan and others},
  journal={arXiv preprint arXiv:2412.00131},
  year={2024}
}

@inproceedings{chenpixart,
  title={PixArt-alpha: Fast Training of Diffusion Transformer for Photorealistic Text-to-Image Synthesis},
  author={Chen, Junsong and Jincheng, YU and Chongjian, GE and Yao, Lewei and Xie, Enze and Wang, Zhongdao and Kwok, James and Luo, Ping and Lu, Huchuan and Li, Zhenguo},
  booktitle={The Twelfth International Conference on Learning Representations}
}

@article{shao2024deepseekmath,
  title={Deepseekmath: Pushing the limits of mathematical reasoning in open language models},
  author={Shao, Zhihong and Wang, Peiyi and Zhu, Qihao and Xu, Runxin and Song, Junxiao and Bi, Xiao and Zhang, Haowei and Zhang, Mingchuan and Li, YK and others},
  journal={arXiv preprint arXiv:2402.03300},
  year={2024}
}

@article{liu2025flow,
  title={Flow-grpo: Training flow matching models via online rl},
  author={Liu, Jie and Liu, Gongye and Liang, Jiajun and Li, Yangguang and Liu, Jiaheng and Wang, Xintao and Wan, Pengfei and Zhang, Di and Ouyang, Wanli},
  journal={arXiv preprint arXiv:2505.05470},
  year={2025}
}

@article{xue2025dancegrpo,
  title={DanceGRPO: Unleashing GRPO on Visual Generation},
  author={Xue, Zeyue and Wu, Jie and Gao, Yu and Kong, Fangyuan and Zhu, Lingting and Chen, Mengzhao and Liu, Zhiheng and Liu, Wei and Guo, Qiushan and Huang, Weilin and others},
  journal={arXiv preprint arXiv:2505.07818},
  year={2025}
}

@article{li2025branchgrpo,
  title={Branchgrpo: Stable and efficient grpo with structured branching in diffusion models},
  author={Li, Yuming and Wang, Yikai and Zhu, Yuying and Zhao, Zhongyu and Lu, Ming and She, Qi and Zhang, Shanghang},
  journal={arXiv preprint arXiv:2509.06040},
  year={2025}
}

@article{li2025mixgrpo,
  title={Mixgrpo: Unlocking flow-based grpo efficiency with mixed ode-sde},
  author={Li, Junzhe and Cui, Yutao and Huang, Tao and Ma, Yinping and Fan, Chun and Yang, Miles and Zhong, Zhao},
  journal={arXiv preprint arXiv:2507.21802},
  year={2025}
}

@article{he2025tempflow,
  title={Tempflow-grpo: When timing matters for grpo in flow models},
  author={He, Xiaoxuan and Fu, Siming and Zhao, Yuke and Li, Wanli and Yang, Jian and Yin, Dacheng and Rao, Fengyun and Zhang, Bo},
  journal={arXiv preprint arXiv:2508.04324},
  year={2025}
}

@article{zheng2025diffusionnft,
  title={Diffusionnft: Online diffusion reinforcement with forward process},
  author={Zheng, Kaiwen and Chen, Huayu and Ye, Haotian and Wang, Haoxiang and Zhang, Qinsheng and Jiang, Kai and Su, Hang and Ermon, Stefano and Zhu, Jun and Liu, Ming-Yu},
  journal={arXiv preprint arXiv:2509.16117},
  year={2025}
}

@article{li2025uniworld,
  title={Uniworld-V2: Reinforce Image Editing with Diffusion Negative-aware Finetuning and MLLM Implicit Feedback},
  author={Li, Zongjian and Liu, Zheyuan and Zhang, Qihui and Lin, Bin and Yuan, Shenghai and Yan, Zhiyuan and Ye, Yang and Yu, Wangbo and Niu, Yuwei and Yuan, Li},
  journal={arXiv preprint arXiv:2510.16888},
  year={2025}
}

@article{xue2025advantage,
  title={Advantage weighted matching: Aligning rl with pretraining in diffusion models},
  author={Xue, Shuchen and Ge, Chongjian and Zhang, Shilong and Li, Yichen and Ma, Zhi-Ming},
  journal={arXiv preprint arXiv:2509.25050},
  year={2025}
}

@article{fu2025dynamic,
  title={Dynamic-TreeRPO: Breaking the Independent Trajectory Bottleneck with Structured Sampling},
  author={Fu, Xiaolong and Ma, Lichen and Guo, Zipeng and Zhou, Gaojing and Wang, Chongxiao and Dong, ShiPing and Zhou, Shizhe and Liu, Ximan and Fu, Jingling and Sin, Tan Lit and others},
  journal={arXiv preprint arXiv:2509.23352},
  year={2025}
}

@article{chen2025skyreels,
  title={Skyreels-v2: Infinite-length film generative model},
  author={Chen, Guibin and Lin, Dixuan and Yang, Jiangping and Lin, Chunze and Zhu, Junchen and Fan, Mingyuan and Zhang, Hao and Chen, Sheng and Chen, Zheng and Ma, Chengcheng and others},
  journal={arXiv preprint arXiv:2504.13074},
  year={2025}
}

@article{mao2025omni,
  title={Omni-effects: Unified and spatially-controllable visual effects generation},
  author={Mao, Fangyuan and Hao, Aiming and Chen, Jintao and Liu, Dongxia and Feng, Xiaokun and Zhu, Jiashu and Wu, Meiqi and Chen, Chubin and Wu, Jiahong and Chu, Xiangxiang},
  journal={arXiv preprint arXiv:2508.07981},
  year={2025}
}

@inproceedings{yang2024hi3d,
  title={Hi3d: Pursuing high-resolution image-to-3d generation with video diffusion models},
  author={Yang, Haibo and Chen, Yang and Pan, Yingwei and Yao, Ting and Chen, Zhineng and Ngo, Chong-Wah and Mei, Tao},
  booktitle={Proceedings of the 32nd ACM International Conference on Multimedia},
  pages={6870--6879},
  year={2024}
}

@inproceedings{hu2024animate,
  title={Animate anyone: Consistent and controllable image-to-video synthesis for character animation},
  author={Hu, Li},
  booktitle={Proceedings of the IEEE/CVF Conference on Computer Vision and Pattern Recognition},
  pages={8153--8163},
  year={2024}
}

@inproceedings{he2020momentum,
  title={Momentum contrast for unsupervised visual representation learning},
  author={He, Kaiming and Fan, Haoqi and Wu, Yuxin and Xie, Saining and Girshick, Ross},
  booktitle={Proceedings of the IEEE/CVF conference on computer vision and pattern recognition},
  pages={9729--9738},
  year={2020}
}

@misc{openai2024sora,
  author       = {OpenAI},
  title        = {Sora: Creating video from text},
  year         = {2024},
  note         = {Accessed: 2025-11-07}
}

@article{wang2024vidprom,
  title={Vidprom: A million-scale real prompt-gallery dataset for text-to-video diffusion models},
  author={Wang, Wenhao and Yang, Yi},
  journal={Advances in Neural Information Processing Systems},
  volume={37},
  pages={65618--65642},
  year={2024}
}

@inproceedings{chen2024panda,
  title={Panda-70m: Captioning 70m videos with multiple cross-modality teachers},
  author={Chen, Tsai-Shien and Siarohin, Aliaksandr and Menapace, Willi and Deyneka, Ekaterina and Chao, Hsiang-wei and Jeon, Byung Eun and Fang, Yuwei and Lee, Hsin-Ying and Ren, Jian and Yang, Ming-Hsuan and others},
  booktitle={Proceedings of the IEEE/CVF Conference on Computer Vision and Pattern Recognition},
  pages={13320--13331},
  year={2024}
}

@misc{google2024veo,
  author       = {Google DeepMind},
  title        = {Veo3: Our state-of-the-art video generation model},
  year         = {2025},
  note         = {Accessed: 2025-11-07}
}

@article{gao2025seedance,
  title={Seedance 1.0: Exploring the Boundaries of Video Generation Models},
  author={Gao, Yu and Guo, Haoyuan and Hoang, Tuyen and Huang, Weilin and Jiang, Lu and Kong, Fangyuan and Li, Huixia and Li, Jiashi and Li, Liang and Li, Xiaojie and others},
  journal={arXiv preprint arXiv:2506.09113},
  year={2025}
}

@article{voleti2022mcvd,
  title={Mcvd-masked conditional video diffusion for prediction, generation, and interpolation},
  author={Voleti, Vikram and Jolicoeur-Martineau, Alexia and Pal, Chris},
  journal={Advances in neural information processing systems},
  volume={35},
  pages={23371--23385},
  year={2022}
}

@inproceedings{chen2023seine,
  title={Seine: Short-to-long video diffusion model for generative transition and prediction},
  author={Chen, Xinyuan and Wang, Yaohui and Zhang, Lingjun and Zhuang, Shaobin and Ma, Xin and Yu, Jiashuo and Wang, Yali and Lin, Dahua and Qiao, Yu and Liu, Ziwei},
  booktitle={The Twelfth International Conference on Learning Representations},
  year={2023}
}

@inproceedings{radford2021learning,
  title={Learning transferable visual models from natural language supervision},
  author={Radford, Alec and Kim, Jong Wook and Hallacy, Chris and Ramesh, Aditya and Goh, Gabriel and Agarwal, Sandhini and Sastry, Girish and Askell, Amanda and Mishkin, Pamela and Clark, Jack and others},
  booktitle={International conference on machine learning},
  pages={8748--8763},
  year={2021},
  organization={PmLR}
}

@article{zhang2023i2vgen,
  title={I2vgen-xl: High-quality image-to-video synthesis via cascaded diffusion models},
  author={Zhang, Shiwei and Wang, Jiayu and Zhang, Yingya and Zhao, Kang and Yuan, Hangjie and Qin, Zhiwu and Wang, Xiang and Zhao, Deli and Zhou, Jingren},
  journal={arXiv preprint arXiv:2311.04145},
  year={2023}
}

@article{chen2023videocrafter1,
  title={Videocrafter1: Open diffusion models for high-quality video generation},
  author={Chen, Haoxin and Xia, Menghan and He, Yingqing and Zhang, Yong and Cun, Xiaodong and Yang, Shaoshu and Xing, Jinbo and Liu, Yaofang and Chen, Qifeng and Wang, Xintao and others},
  journal={arXiv preprint arXiv:2310.19512},
  year={2023}
}

@inproceedings{shi2024motion,
  title={Motion-i2v: Consistent and controllable image-to-video generation with explicit motion modeling},
  author={Shi, Xiaoyu and Huang, Zhaoyang and Wang, Fu-Yun and Bian, Weikang and Li, Dasong and Zhang, Yi and Zhang, Manyuan and Cheung, Ka Chun and See, Simon and Qin, Hongwei and others},
  booktitle={ACM SIGGRAPH 2024 Conference Papers},
  pages={1--11},
  year={2024}
}

@inproceedings{zeng2024make,
  title={Make pixels dance: High-dynamic video generation},
  author={Zeng, Yan and Wei, Guoqiang and Zheng, Jiani and Zou, Jiaxin and Wei, Yang and Zhang, Yuchen and Li, Hang},
  booktitle={Proceedings of the IEEE/CVF Conference on Computer Vision and Pattern Recognition},
  pages={8850--8860},
  year={2024}
}

@article{wang2023videocomposer,
  title={Videocomposer: Compositional video synthesis with motion controllability},
  author={Wang, Xiang and Yuan, Hangjie and Zhang, Shiwei and Chen, Dayou and Wang, Jiuniu and Zhang, Yingya and Shen, Yujun and Zhao, Deli and Zhou, Jingren},
  journal={Advances in Neural Information Processing Systems},
  volume={36},
  pages={7594--7611},
  year={2023}
}

@inproceedings{xing2024dynamicrafter,
  title={Dynamicrafter: Animating open-domain images with video diffusion priors},
  author={Xing, Jinbo and Xia, Menghan and Zhang, Yong and Chen, Haoxin and Yu, Wangbo and Liu, Hanyuan and Liu, Gongye and Wang, Xintao and Shan, Ying and Wong, Tien-Tsin},
  booktitle={European Conference on Computer Vision},
  pages={399--417},
  year={2024},
  organization={Springer}
}

@inproceedings{tian2025extrapolating,
  title={Extrapolating and Decoupling Image-to-Video Generation Models: Motion Modeling is Easier Than You Think},
  author={Tian, Jie and Qu, Xiaoye and Lu, Zhenyi and Wei, Wei and Liu, Sichen and Cheng, Yu},
  booktitle={Proceedings of the Computer Vision and Pattern Recognition Conference},
  pages={12512--12521},
  year={2025}
}

@inproceedings{guo2024i2v,
  title={I2v-adapter: A general image-to-video adapter for diffusion models},
  author={Guo, Xun and Zheng, Mingwu and Hou, Liang and Gao, Yuan and Deng, Yufan and Wan, Pengfei and Zhang, Di and Liu, Yufan and Hu, Weiming and Zha, Zhengjun and others},
  booktitle={ACM SIGGRAPH 2024 Conference Papers},
  pages={1--12},
  year={2024}
}

@article{xu2023imagereward,
  title={Imagereward: Learning and evaluating human preferences for text-to-image generation},
  author={Xu, Jiazheng and Liu, Xiao and Wu, Yuchen and Tong, Yuxuan and Li, Qinkai and Ding, Ming and Tang, Jie and Dong, Yuxiao},
  journal={Advances in Neural Information Processing Systems},
  volume={36},
  pages={15903--15935},
  year={2023}
}

@article{shen2025directly,
  title={Directly aligning the full diffusion trajectory with fine-grained human preference},
  author={Shen, Xiangwei and Li, Zhimin and Yang, Zhantao and Zhang, Shiyi and Zhang, Yingfang and Li, Donghao and Wang, Chunyu and Lu, Qinglin and Tang, Yansong},
  journal={arXiv preprint arXiv:2509.06942},
  year={2025}
}

@article{prabhudesai2023aligning,
  title={Aligning text-to-image diffusion models with reward backpropagation},
  author={Prabhudesai, Mihir and Goyal, Anirudh and Pathak, Deepak and Fragkiadaki, Katerina},
  year={2023}
}

@article{clark2023directly,
  title={Directly fine-tuning diffusion models on differentiable rewards},
  author={Clark, Kevin and Vicol, Paul and Swersky, Kevin and Fleet, David J},
  journal={arXiv preprint arXiv:2309.17400},
  year={2023}
}

@article{prabhudesai2024video,
  title={Video diffusion alignment via reward gradients},
  author={Prabhudesai, Mihir and Mendonca, Russell and Qin, Zheyang and Fragkiadaki, Katerina and Pathak, Deepak},
  journal={arXiv preprint arXiv:2407.08737},
  year={2024}
}

@article{rafailov2023direct,
  title={Direct preference optimization: Your language model is secretly a reward model},
  author={Rafailov, Rafael and Sharma, Archit and Mitchell, Eric and Manning, Christopher D and Ermon, Stefano and Finn, Chelsea},
  journal={Advances in neural information processing systems},
  volume={36},
  pages={53728--53741},
  year={2023}
}

@inproceedings{wallace2024diffusion,
  title={Diffusion model alignment using direct preference optimization},
  author={Wallace, Bram and Dang, Meihua and Rafailov, Rafael and Zhou, Linqi and Lou, Aaron and Purushwalkam, Senthil and Ermon, Stefano and Xiong, Caiming and Joty, Shafiq and Naik, Nikhil},
  booktitle={Proceedings of the IEEE/CVF Conference on Computer Vision and Pattern Recognition},
  pages={8228--8238},
  year={2024}
}

@inproceedings{liu2025videodpo,
  title={Videodpo: Omni-preference alignment for video diffusion generation},
  author={Liu, Runtao and Wu, Haoyu and Zheng, Ziqiang and Wei, Chen and He, Yingqing and Pi, Renjie and Chen, Qifeng},
  booktitle={Proceedings of the Computer Vision and Pattern Recognition Conference},
  pages={8009--8019},
  year={2025}
}

@inproceedings{yang2024using,
  title={Using human feedback to fine-tune diffusion models without any reward model},
  author={Yang, Kai and Tao, Jian and Lyu, Jiafei and Ge, Chunjiang and Chen, Jiaxin and Shen, Weihan and Zhu, Xiaolong and Li, Xiu},
  booktitle={Proceedings of the IEEE/CVF Conference on Computer Vision and Pattern Recognition},
  pages={8941--8951},
  year={2024}
}

@article{yuan2024self,
  title={Self-play fine-tuning of diffusion models for text-to-image generation},
  author={Yuan, Huizhuo and Chen, Zixiang and Ji, Kaixuan and Gu, Quanquan},
  journal={Advances in Neural Information Processing Systems},
  volume={37},
  pages={73366--73398},
  year={2024}
}

@article{zhang2024onlinevpo,
  title={Onlinevpo: Align video diffusion model with online video-centric preference optimization},
  author={Zhang, Jiacheng and Wu, Jie and Chen, Weifeng and Ji, Yatai and Xiao, Xuefeng and Huang, Weilin and Han, Kai},
  journal={arXiv preprint arXiv:2412.15159},
  year={2024}
}

@article{furuta2024improving,
  title={Improving dynamic object interactions in text-to-video generation with ai feedback},
  author={Furuta, Hiroki and Zen, Heiga and Schuurmans, Dale and Faust, Aleksandra and Matsuo, Yutaka and Liang, Percy and Yang, Sherry},
  journal={arXiv preprint arXiv:2412.02617},
  year={2024}
}

@inproceedings{liang2025aesthetic,
  title={Aesthetic post-training diffusion models from generic preferences with step-by-step preference optimization},
  author={Liang, Zhanhao and Yuan, Yuhui and Gu, Shuyang and Chen, Bohan and Hang, Tiankai and Cheng, Mingxi and Li, Ji and Zheng, Liang},
  booktitle={Proceedings of the Computer Vision and Pattern Recognition Conference},
  pages={13199--13208},
  year={2025}
}

@misc{SkyReelsV1,
  author = {SkyReels-AI},
  title = {Skyreels V1: Human-Centric Video Foundation Model},
  year = {2025},
  publisher = {GitHub},
  journal = {GitHub repository},
  howpublished = {\url{https://github.com/SkyworkAI/SkyReels-V1}}
}

@inproceedings{ma2025hpsv3,
  title={Hpsv3: Towards wide-spectrum human preference score},
  author={Ma, Yuhang and Wu, Xiaoshi and Sun, Keqiang and Li, Hongsheng},
  booktitle={Proceedings of the IEEE/CVF International Conference on Computer Vision},
  pages={15086--15095},
  year={2025}
}

@article{wu2025hunyuanvideo,
  title={HunyuanVideo 1.5 Technical Report},
  author={Wu, Bing and Zou, Chang and Li, Changlin and Huang, Duojun and Yang, Fang and Tan, Hao and Peng, Jack and Wu, Jianbing and Xiong, Jiangfeng and Jiang, Jie and others},
  journal={arXiv preprint arXiv:2511.18870},
  year={2025}
}

@article{wu2025q,
  title={Q-Save: Towards Scoring and Attribution for Generated Video Evaluation},
  author={Wu, Xiele and Zhang, Zicheng and Chen, Mingtao and Liu, Yixian and Liu, Yiming and Wang, Shushi and Hu, Zhichao and Liu, Yuhong and Zhai, Guangtao and Liu, Xiaohong},
  journal={arXiv preprint arXiv:2511.18825},
  year={2025}
}

@article{du2025reg,
  title={Reg-DPO: SFT-Regularized Direct Preference Optimization with GT-Pair for Improving Video Generation},
  author={Du, Jie and Gong, Xinyu and Tan, Qingshan and Li, Wen and Cheng, Yangming and Wang, Weitao and Zhan, Chenlu and Wu, Suhui and Zhang, Hao and Zhang, Jun},
  journal={arXiv preprint arXiv:2511.01450},
  year={2025}
}

@article{huang2025vbench++,
  title={Vbench++: Comprehensive and versatile benchmark suite for video generative models},
  author={Huang, Ziqi and Zhang, Fan and Xu, Xiaojie and He, Yinan and Yu, Jiashuo and Dong, Ziyue and Ma, Qianli and Chanpaisit, Nattapol and Si, Chenyang and Jiang, Yuming and others},
  journal={IEEE Transactions on Pattern Analysis and Machine Intelligence},
  year={2025},
  publisher={IEEE}
}

@inproceedings{zhang2018unreasonable,
  title={The unreasonable effectiveness of deep features as a perceptual metric},
  author={Zhang, Richard and Isola, Phillip and Efros, Alexei A and Shechtman, Eli and Wang, Oliver},
  booktitle={Proceedings of the IEEE conference on computer vision and pattern recognition},
  pages={586--595},
  year={2018}
}
\bibliographystyle{omniweaving_arxiv}
\clearpage
\appendix
\section{Appendix: Human Evaluations}
To further validate the effectiveness of the TAGRPO method, we conducted a rigorous human evaluation study to assess the quality of our image-to-video (I2V) generation.

\subsection{Evaluation Dimensions and Scoring Criteria}
We developed a multi-dimensional human evaluation framework based on industry standards and product requirements. The evaluation encompasses five primary dimensions: Image-Video Consistency (IV), Text-Video Consistency (TV), Static Quality (SQ), Dynamic Quality (DQ), and Overall Assessment (OA). Each dimension employs a 5-point Likert scale (1: Poor to 5: Excellent) to quantify performance.

\subsubsection{Image-Video Consistency}
This dimension evaluates how well the generated video maintains visual fidelity to the reference image across temporal frames. The score criteria is as follows.
\begin{itemize}
    \item \textbf{5 (Excellent):} Near-perfect consistency ($\ge$90\%) between video content and reference image.
    \item \textbf{4 (Good):} Substantial consistency ($\ge$70\%) with minor deviations.
    \item \textbf{3 (Average):} Moderate consistency ($\ge$50\%) with noticeable differences.
    \item \textbf{2 (Below Average):} Significant inconsistency ($<$50\%) from reference.
    \item \textbf{1 (Poor):} Complete inconsistency ($<$20\%) with reference image.
\end{itemize}

\subsubsection{Text-Video Consistency}
This dimension measures the alignment between the generated video and the provided textual instructions. The score criteria is as follows.
\begin{itemize}
    \item \textbf{5 (Excellent):} Near-complete adherence ($\ge$80\%) to text instructions.
    \item \textbf{4 (Good):} Majority compliance ($\ge$50\%) with text guidance.
    \item \textbf{3 (Average):} Partial compliance ($<$50\%) with instructions.
    \item \textbf{2 (Below Average):} Minimal adherence ($<$25\%) to text prompts.
    \item \textbf{1 (Poor):} Complete non-compliance or motion failure.
\end{itemize}

\subsubsection{Static Quality}
This dimension evaluates the visual richness, clarity, and aesthetic appeal of individual frames independent of motion.  The score criteria is as follows.
\begin{itemize}
    \item \textbf{5 (Excellent):} Exceptional detail richness with high clarity and aesthetic appeal.
    \item \textbf{4 (Good):} Substantial detail visibility with good clarity and aesthetics.
    \item \textbf{3 (Average):} Moderate detail level with acceptable clarity and aesthetics.
    \item \textbf{2 (Below Average):} Limited detail presence with blurriness and average aesthetics.
    \item \textbf{1 (Poor):} Minimal detail with severe artifacts and poor aesthetics.
\end{itemize}

\subsubsection{Dynamic Quality}
This dimension assesses the naturalness, fluidity, and realism of motion in generated videos.  The score criteria is as follows.
\begin{itemize}
    \item \textbf{5 (Excellent):} Natural, fluid motion approaching professional quality.
    \item \textbf{4 (Good):} Generally reasonable motion with minor artifacts.
    \item \textbf{3 (Average):} Basic motion naturalness with noticeable distortions.
    \item \textbf{2 (Below Average):} Significant motion abnormalities and distortions.
    \item \textbf{1 (Poor):} Severe motion artifacts and unnatural character movement.
\end{itemize}

\subsubsection{Overall Assessment}
This holistic dimension provides an integrated evaluation of the total generation quality.
\begin{itemize}
    \item \textbf{5 (Excellent):} Strong feature consistency, fluid motion, and strict text compliance.
    \item \textbf{4 (Good):} Minor motion artifacts but maintains substantial consistency and compliance.
    \item \textbf{3 (Average):} Acceptable quality with basic compliance and minimal visual artifacts.
    \item \textbf{2 (Below Average):} Limited reference matching with significant motion abnormalities.
    \item \textbf{1 (Poor):} Severe quality issues, including motion failure and structural distortions.
\end{itemize}

\subsection{Evaluation Methodology and Participants}
To ensure the practical applicability of these scores, we recruited 18 domain experts to perform the assessment. 
These participants possess extensive professional experience in computer graphics and computer vision. By applying the multi-dimensional criteria defined above, the expert panel provided a rigorous quantitative and qualitative analysis, ensuring that the evaluation of TAGRPO reflects high-tier standards for I2V generations.
The results are summarized in Table \ref{tab:user_study}. Our method was evaluated across two state-of-the-art base models: Wan 2.2 and HunyuanVideo 1.5, using HPSv3 and Q-Save reward configurations for TAGRPO.
The empirical results demonstrate that TAGRPO significantly enhanced the performance of leading I2V diffusion models. By introducing trajectory-wise alignment and leveraging high-fidelity reward models like Q-Save, our framework consistently outperformed base models across nearly all critical dimensions, establishing TAGRPO as a scalable and effective strategy for post-training image-to-video generative models.

\begin{table}[t]
\centering
\caption{Quantitative comparisons in the human evaluation of our benchmark. The results demonstrate that our method delivered superior performance in terms of Image-Video Consistency (IV), Text-Video Consistency (TV), Static Quality (SQ), Dynamic Quality (DQ), and Overall Assessment (OA).}
{
\linespread{1.2}
\setlength\tabcolsep{10pt}
\small
{%
\begin{tabular}{lcccccc}
\toprule
\textbf{Method} & \textbf{IV} & \textbf{TV} & \textbf{SQ} & \textbf{DQ}& \textbf{OA}\\
\midrule
Wan 2.2& 3.60 & 3.41 & 3.22  & 3.28& 3.20 \\
\textbf{+TAGRPO (HPSv3)}& 3.61 & 3.46 & 3.30 & 3.39  & 3.28\\
\textbf{+TAGRPO (Q-Save)}& 3.73 & 3.55 & 3.39 & 3.40 & 3.35 \\
\midrule
HunyuanVideo 1.5& 3.23 & 3.73 & 2.84  & 2.60& 2.59 \\
\textbf{+TAGRPO (Q-Save)}& 3.29 & 2.73 & 2.90 & 2.63 & 2.62 \\
\bottomrule
\end{tabular}
}
}
\label{tab:user_study}
\end{table}


\section{More Experimental Results}

\begin{figure}[t]
  \centering
  
  \begin{minipage}[t]{0.48\textwidth}
    \centering
    \includegraphics[width=\textwidth]{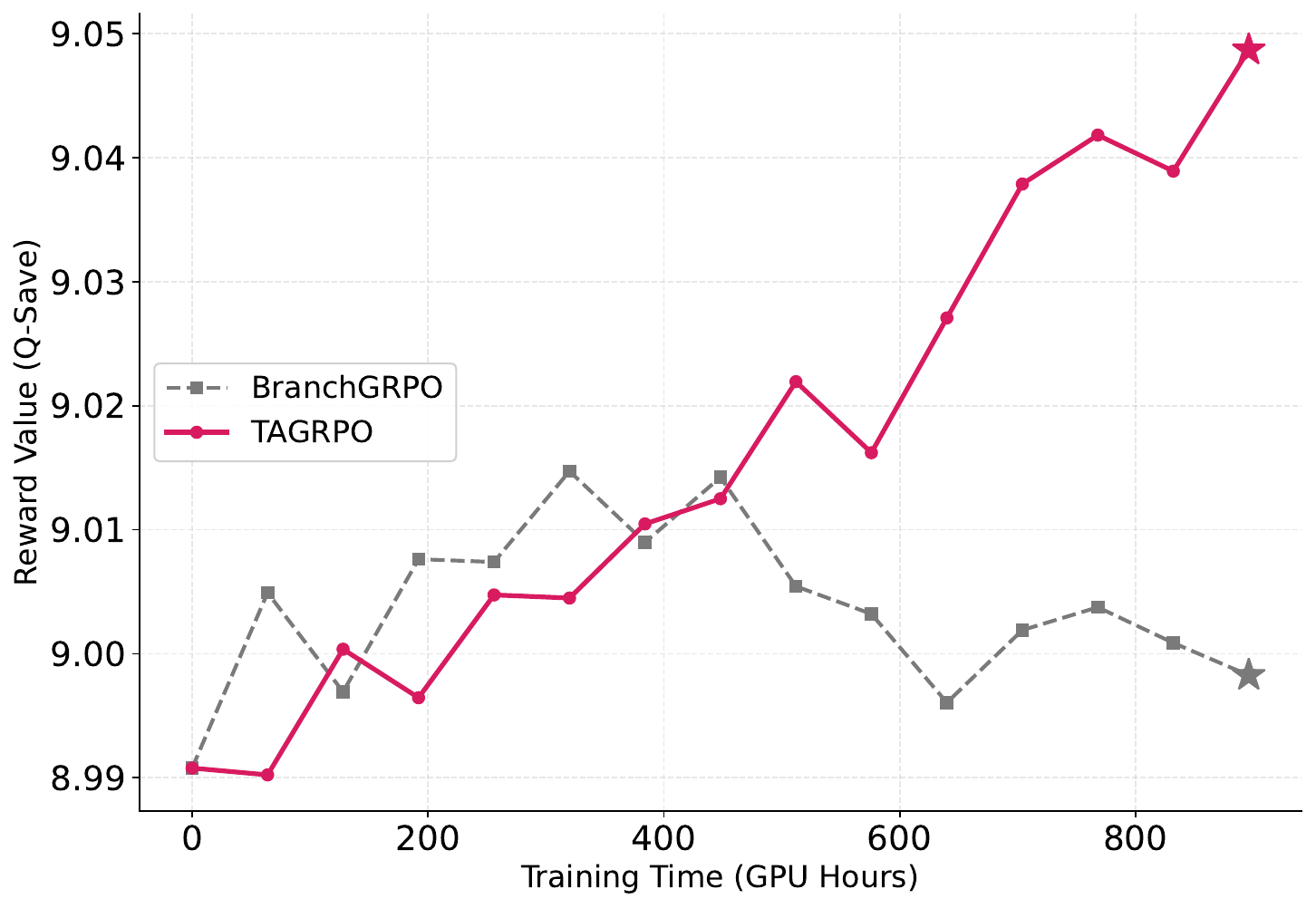}
    \caption{Performance comparison between TAGRPO and BranchGRPO on Wan 2.2. TAGRPO achieved superior reward convergence in I2V generation by leveraging shared structural context via $\mathcal{J}_{\text{align}}$.}
    \label{fig:ablationbranchgrpo}
  \end{minipage}%
  \hfill 
  \begin{minipage}[t]{0.48\textwidth}
    \centering
    \includegraphics[width=\textwidth]{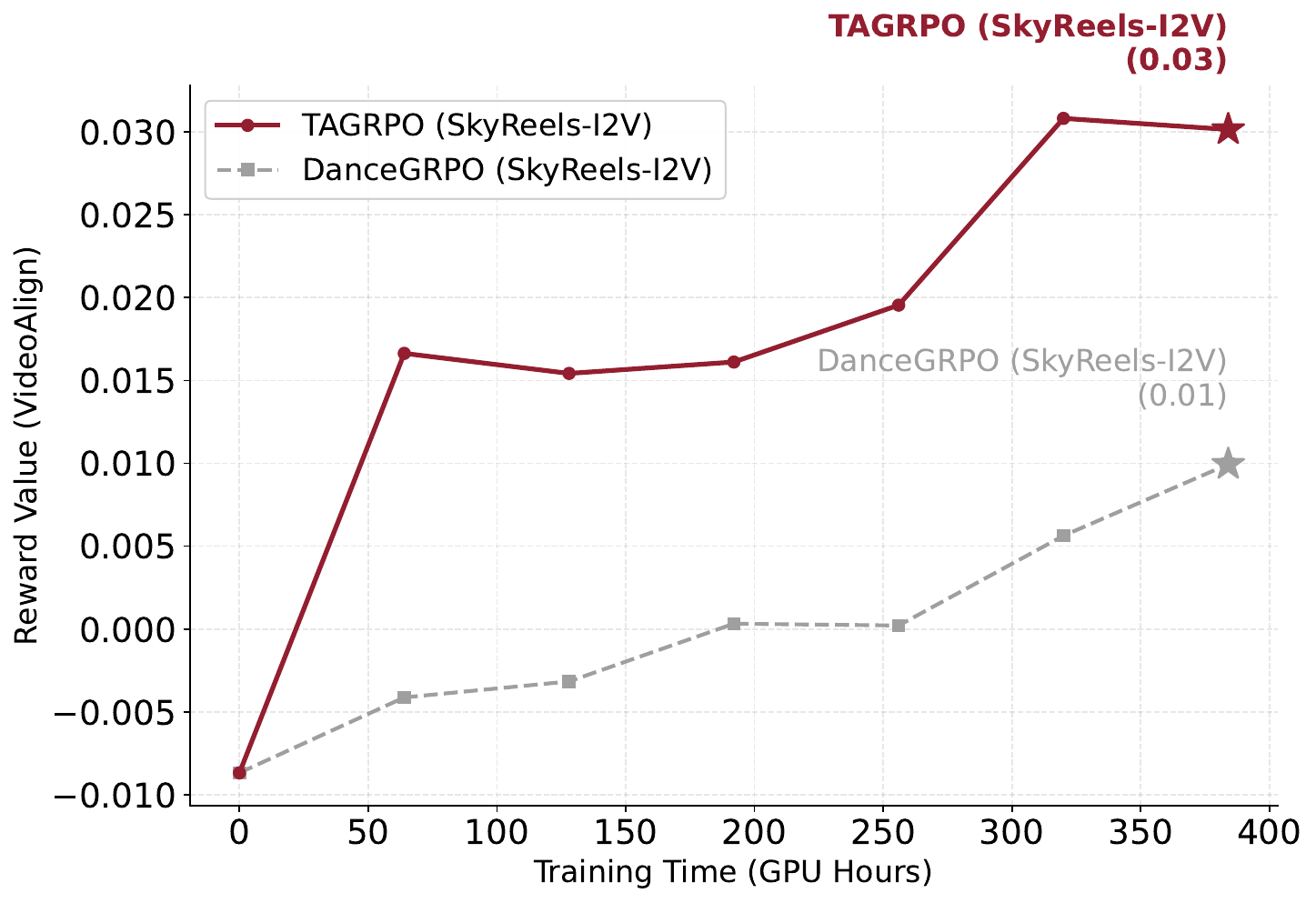}
    \caption{Performance comparison on the SkyReels-I2V base model. While DanceGRPO achieved marginal gains, TAGRPO consistently delivered a higher reward upper bound.}
    \label{fig:ablationskyreels}
  \end{minipage}
  
\end{figure}

\textbf{More Comparisons with BranchGRPO}. In this section, we conducted new experiments comparing TAGRPO with BranchGRPO \cite{li2025branchgrpo}, using Wan 2.2 as the base model. Since BranchGRPO was not developed on I2V tasks, we carefully reproduced it for this setting. As shown in Figure \ref{fig:ablationbranchgrpo}, TAGRPO achieved superior reward convergence. While BranchGRPO focused on sampling efficiency, its independent advantage estimation still struggled with I2V tasks. TAGRPO addressed this via $\mathcal{J}_{\text{align}}$, leveraging shared structural context to provide more discriminative signals. 

\textbf{More Results on SkyReels-I2V}. In this section, we conducted new experiments using another base model, SkyReels-I2V \cite{SkyReelsV1}. As shown in Figure \ref{fig:ablationskyreels}, in this setting, DanceGRPO indeed achieved gains on this simpler base, yet TAGRPO still consistently delivered a higher reward upper bound. These results, combined with our sensitivity analysis (see Section \ref{section:Sensitivity_gamma}), confirm that standard GRPO was sensitive to model-specific tuning and prone to stagnation on high-performance models. In contrast, TAGRPO was highly robust, achieving consistent reward growth across a wide range of hyperparameter configurations and diverse base models. This cross-model success proved that TAGRPO provided robust, discriminative signals necessary to break optimization stagnation where standard methods failed.

\textbf{About Generation Diversity}. In this section, we conducted new experiments to evaluate generation diversity using the V-LPIPS \cite{zhang2018unreasonable} metric, confirming that TAGRPO effectively prevented mode collapse and actually enhanced diversity. Specifically, we generated 5 videos per prompt using a subset of the VBench-I2V \cite{huang2025vbench++} dataset and calculated the average pairwise V-LPIPS (i.e., by averaging the frame-level LPIPS scores). As shown in Table \ref{tab:generation_diversity}, TAGRPO achieved a higher diversity score than the base model.
We attributed this robust diversity preservation to our memory bank mechanism. By continuously exposing the model to diverse historical samples from different training stages, it acted as a natural regularizer that prevented the policy from collapsing into a single, repetitive trajectory pattern.

\textbf{More Results on VBench-I2V}. In this section, we incorporated the standardized VBench-I2V \cite{huang2025vbench++} benchmark into our evaluation, completing the assessment for the ``I2V Subject" metric. As shown in Table \ref{tab:Vbench-i2v}, TAGRPO achieved a measurable improvement over the strong base model.

\section{More Discussions}
\textbf{Detailed Comparisons with DPO}. The advantage function of our method distinguishes it from DPO \cite{rafailov2023direct} in several key structural aspects:
\begin{itemize}
    \item \textit{Dynamic Update Scales}: While DPO relies on static binary preferences (essentially treating pairs as fixed +1/-1 targets regardless of the actual reward margin), the gradient magnitude of $\mathcal{J}_{\text{align}}$ is explicitly and dynamically modulated by the continuous advantage values ($\hat{A}^{+}$ and $\hat{A}^{-}$) of the extreme samples (Eq. \ref{eq:align_loss}). This ensures that the alignment strength is strictly proportional to how much better or worse a trajectory is relative to the group—a dynamic scaling mechanism driven entirely by the advantage function and fundamentally absent in DPO.
    \item \textit{TAGRPO preserves $\mathcal{J}_{\text{GRPO}}$}: As shown in Eq. 19, our final objective is $\mathcal{J}_{\text{TAGRPO}} = \mathcal{J}_{\text{GRPO}} + \gamma \mathcal{J}_{\text{align}}$. Thus, the per-sample advantages are still fully retained in $\mathcal{J}_{\text{GRPO}}$ for the global policy update in our TAGRPO framework.
    \item \textit{Inter-Sample Interactions via Top-/Bottom-$k$ Subsets}: While our current implementation selects the single best and worst samples in each group for computational efficiency, our mathematical formulation naturally generalizes to broader inter-sample interactions by utilizing top-$k$ and bottom-$k$ trajectories for guidance, which can simultaneously process and dynamically weight multiple trajectories based on their continuous advantage scores. 
    \item \textit{PPO-Style Clipping}: Unlike DPO, which directly optimizes log-likelihood ratios over static preference pairs, our $\mathcal{J}_{\text{align}}$ (Eqs. \ref{eq:align_loss}) operates on PPO-style clipped importance ratios ($r^i_t(\theta)^{+}$ and $r^i_t(\theta)^{-}$). This inherently restricts the policy update, preserving the reinforcement learning policy optimization framework rather than degrading to a DPO-like objective.
\end{itemize}

\textbf{Regarding Using Extrema in $\mathcal{J}_{\text{align}}$}. We understand that using extrema for advantage estimation is an unconventional design choice, which may raise concerns about the training instability of our proposed TAGRPO. However, our framework maintains stability through both I2V-specific properties and explicit mathematical safeguards as follows.
\begin{itemize}
    \item \textit{Compressed reward distribution in I2V}: Unlike general RL settings with high reward variance, I2V trajectories conditioned on the same image exhibit naturally compressed reward distributions. The ``best" and ``worst" trajectories within a local group are often separated by subtle differences in motion dynamics rather than massive structural shifts. In this specific regime, using extrema safely amplifies meaningful optimization signals without introducing the instability that would occur in high-variance environments.
    \item \textit{Mathematical safeguards restrict updates}: Even if a local extremum produces a larger relative advantage, $\mathcal{J}_{align}$ strictly applies the PPO-style clipping mechanism to the cross-sample importance ratios (Eq. \ref{eq:align_loss}). This imposes a hard mathematical bound on the policy update step, preventing any single extreme trajectory from causing drastic policy shifts.
    \item \textit{Empirical evidence of stability}: Our comprehensive sensitivity analyses on the alignment weight ($\gamma \in \{1.0, 3.0, 5.0\}$) and memory bank size ($N \in \{4, 8, 16\}$)—detailed in Section \ref{section:memorybank_design} and Section \ref{section:Sensitivity_gamma}—demonstrate remarkably stable training dynamics. As shown in the previously provided figures, all configurations converge smoothly without oscillation, even at $\gamma=5.0$ (5x the default value). Furthermore, this stability consistently holds across a diverse set of state-of-the-art architectures and tasks, including Wan2.2-I2V/T2V-A14B, HunyuanVideo-1.5-I2V/T2V, SkyReels-I2V, and Wan2.2-TI2V-5B. This broad success suggests that the extrema-based design is significantly more robust than intuition might suggest.
\end{itemize}

\begin{table}[t]
  \centering
  
  \begin{minipage}[t]{0.48\textwidth}
    \centering
    \caption{Quantitative evaluation of generation diversity. Measured by average pairwise V-LPIPS on a subset of VBench-I2V, TAGRPO successfully prevented mode collapse and enhanced the diversity of the base model.}
    \label{tab:generation_diversity}
    {
    \linespread{1.2}
    \setlength\tabcolsep{10pt}
    \small
    \begin{tabular}{lc}
    \toprule
    \textbf{Model} & \textbf{V-LPIPS} $\uparrow$ \\
    \midrule
    Wan 2.2& 0.3824  \\
    \textbf{+TAGRPO}& 0.4086\\
    \bottomrule
    \end{tabular}
    }
  \end{minipage}%
  \hfill 
  \begin{minipage}[t]{0.48\textwidth}
    \centering
    \caption{Quantitative evaluation on the VBench-I2V benchmark. TAGRPO achieved a measurable improvement over the Wan 2.2 base model on the I2V Subject metric.}
    \label{tab:Vbench-i2v}
    {
    \linespread{1.2}
    \setlength\tabcolsep{10pt}
    \small
    \begin{tabular}{lc}
    \toprule
    \textbf{Model} & \textbf{I2V Subject} $\uparrow$ \\
    \midrule
    Wan 2.2& 0.9653  \\
    \textbf{+TAGRPO}& 0.9667\\
    \bottomrule
    \end{tabular}
    }
  \end{minipage}

\end{table}

\textbf{More Analyses on TAGRPO's Motivations}. We observe that standard GRPO struggles with I2V due to strong structural constraints, a limitation that TAGRPO actively resolves through cross-sample interaction. Because the conditioning image imposes a strong structural prior, outputs within a group share highly similar structures. This drastically reduces reward variance and yields weak advantage signals $\hat{A}^{i}$, causing standard GRPO (which treats samples independently) to neglect subtle motion dynamics. TAGRPO addresses this via $\mathcal{J}_{align}$. Rather than relying on isolated per-sample scaling, it explicitly contrasts current trajectories against the best and worst samples within the group. This leverages collective group information to amplify the optimization signal, providing robust directional gradients even when absolute reward differences are minimal.

\textbf{Regarding Distribution Shift in the Memory Bank and the Role of KL Divergence}. We clarify that the distribution shift from the memory bank is inherently minimized by our FIFO strategy, while the KL divergence serves a completely independent role. 
Since the policy's weights update gradually, our FIFO mechanism ensures the stored latents remain extremely close to the current policy, making any distribution ``lag" practically negligible. Besides, rather than treating all historical latents as absolute regression targets, $\mathcal{J}_{align}$ selectively extracts only the top- and bottom-ranked samples to serve strictly as relative trajectory anchors. This mechanism adjusts the policy's probabilities to favor optimal generation paths and penalize poor ones. 
As for the KL term, it is completely decoupled from the memory bank, operating solely on current rollouts to anchor the model to the pre-trained reference model.

\section{Limitations}

\begin{figure}[h!]
  \centering
  \includegraphics[width=0.6\textwidth]{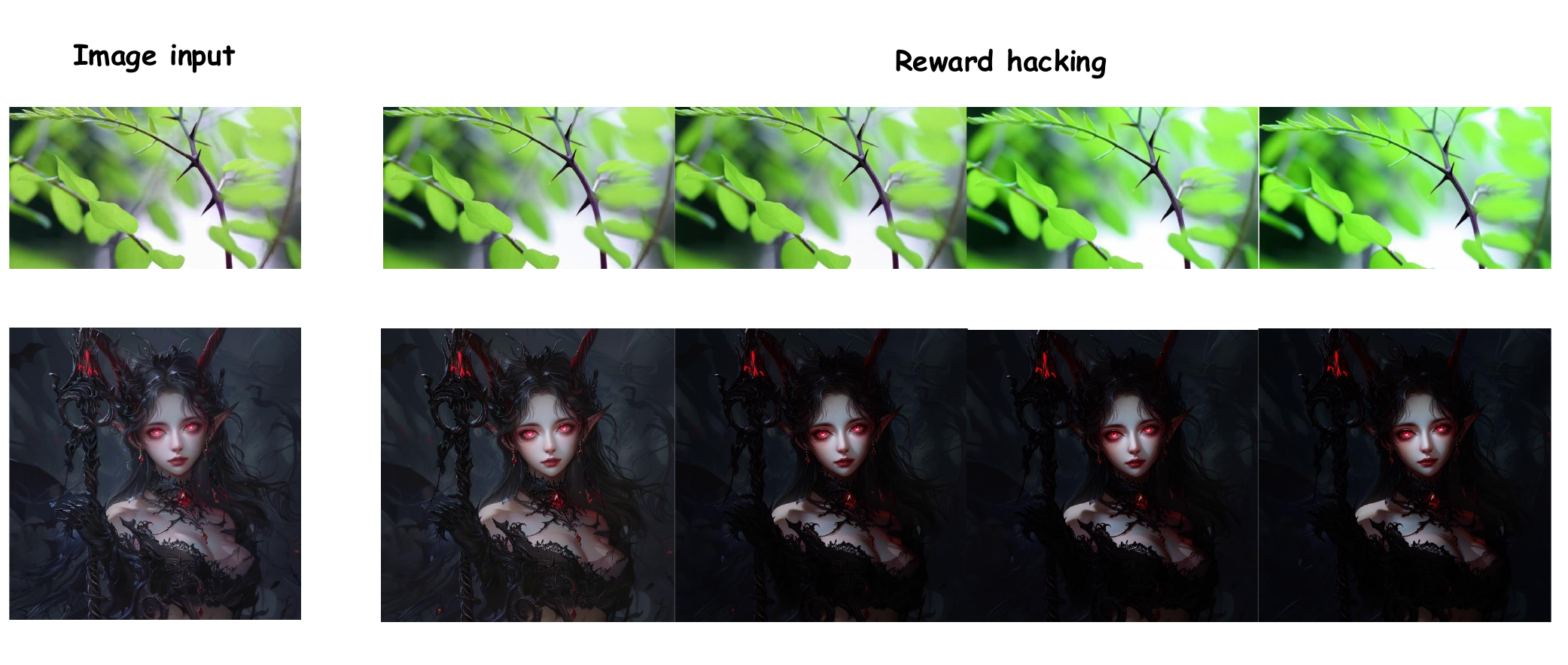}
  \caption{\textbf{Reward hacking phenomenon.} 
  Training with HPSv3 as the sole reward model leads to generated videos exhibiting abnormally high contrast and brightness.}
  \label{fig:reward_hacking}
\end{figure}
As discussed in prior GRPO frameworks~\cite{xue2025dancegrpo,liu2025flow}, reward hacking remains a fundamental challenge when applying reinforcement learning to the visual domain, and our method is not immune to this issue.
As illustrated in Figure~\ref{fig:reward_hacking}, when training with HPSv3 as the sole reward model, the generated videos systematically exhibit higher contrast and brightness compared to the base model outputs.
This behavior arises because HPSv3 tends to assign higher scores to brighter images; consequently, during optimization, the model learns to exploit this bias by producing overly bright or even overexposed videos to maximize reward scores.
Future work may explore multi-reward objectives, adversarial training strategies to detect and mitigate reward hacking, as well as the development of reward models with explicit constraints or regularization to discourage such exploitative behaviors.

\end{document}